\documentclass[runningheads]{llncs}

\usepackage{eccv}

\usepackage{eccvabbrv}
\usepackage{graphicx}
\usepackage{booktabs}
\usepackage{bm}

\usepackage{color,xcolor,xspace}
\usepackage{thm-restate}
\usepackage[font=scriptsize]{caption}

\usepackage[accsupp]{axessibility}  

\usepackage{hyperref}

\usepackage{orcidlink}

\usepackage[capitalize]{cleveref}
\crefname{section}{Sec.}{Secs.}
\Crefname{section}{Section}{Sections}
\Crefname{table}{Table}{Tables}
\crefname{table}{Tab.}{Tabs.}

\begin{document}

\title{VideoMV: Consistent Multi-View Generation Based on Large Video Generative Model}

\titlerunning{VideoMV}

\author{Qi Zuo\inst{1*} \and
Xiaodong Gu\inst{1*} \and \\
Lingteng Qiu\inst{2,1} \and 
Yuan Dong\inst{1} \and 
Zhengyi Zhao\inst{1} \and
Weihao Yuan\inst{1} \and
Rui Peng\inst{4} \and \\
Siyu Zhu\inst{3} \and
Zilong Dong\inst{1} \and
Liefeng Bo\inst{1} \and
Qixing Huang\inst{5} }

\authorrunning{Qi Zuo et al.}

\institute{Institute for Intelligent Computing, Alibaba Group \and
SSE, CUHKSZ \\ \and
Fudan University \\ \and
Peking University \\ \and
The University of Texas at Austin }

\maketitle

\begin{abstract}
Generating multi-view images based on text or single-image prompts is a critical capability for the creation of 3D content. Two fundamental questions on this topic are what data we use for training and how to ensure multi-view consistency. This paper introduces a novel framework that makes fundamental contributions to both questions. Unlike leveraging images from 2D diffusion models for training, we propose a dense consistent multi-view generation model that is fine-tuned from off-the-shelf video generative models. Images from video generative models are more suitable for multi-view generation because the underlying network architecture that generates them employs a temporal module to enforce frame consistency. Moreover, the video data sets used to train these models are abundant and diverse, leading to a reduced train-finetuning domain gap. To enhance multi-view consistency, we introduce a \textit{3D-Aware Denoising Sampling}, which first employs a feed-forward reconstruction module to get an explicit global 3D model, and then adopts a sampling strategy that effectively involves images rendered from the global 3D model into the denoising sampling loop to improve the multi-view consistency of the final images. As a by-product, this module also provides a fast way to create 3D assets represented by 3D Gaussians within a few seconds. Our approach can generate 24 dense views and converges much faster in training than state-of-the-art approaches (4 GPU hours versus many thousand GPU hours) with comparable visual quality and consistency.
By further fine-tuning, our approach outperforms existing state-of-the-art methods in both quantitative metrics and visual effects. Our project page is \textcolor{magenta}{aigc3d.github.io/VideoMV}.

\end{abstract}
\section{Introduction}
\label{sec:intro}

The creation of 3D content plays a crucial role in virtual reality, the game and movie industry, 3D design, and so on. However, the scarcity of large-scale 3D data and the high time consumption of acquiring them pose significant obstacles in learning a strong 3D prior from them for high-quality 3D content creation. To address the data issue, recent advances such as DreamFusion~\cite{poole2022dreamfusion} leverage \textbf{2D generation priors} learned from large-scale image data to optimize different views of the target object. Despite generating realistic views, such approaches suffer from the multi-face janus problem caused by the lack of the underlying 3D model when learning from images generated by 2D diffusion models. Recent approaches, including MVDream~\cite{Shi2023MVDreamMD} and Wonder3D~\cite{Long2023Wonder3DSI}, use the attention layers learned from limited 3D data~\cite{Deitke2022ObjaverseAU} to boost multi-view consistency in the generated images. However, these approaches still present noticeable artifacts in multi-view inconsistency and show limited generalizability. 

We argue that there are two key factors to achieve high-quality and multi-view consistent image generation results. The first is what data and model we use for pre-training. They dictate the type of features being learned, which are important for multi-view consistency. The second factor is how to infer an underlying 3D model, which is the most effective way to enforce multi-view consistency.

This paper introduces VideoMV, a novel approach that makes important contributions to both factors. The key idea of VideoMV is to learn \textbf{video generation priors} from object-central videos. This approach has three key advantages. First, the data scale of object-central videos is large enough to learn strong video generation priors. Second, video generative models have strong attention modules across the frames, which are important for multi-view consistency~\cite{Shi2023MVDreamMD,Long2023Wonder3DSI}.  
Third, frames in a video are projected from different views of a 3D scene, such that these frames follow an underlying 3D model and present continuous and gradual changes, making it easier to learn cross-frame patterns that enforce multi-view consistency.
VideoMV introduces a novel approach to fine-tune a pre-trained video generative model for dense multi-view generation. It only uses a small high-quality 3D dataset. We show how to connect multi-view images of objects with object-centric videos by adding the camera embedding as a residual to the time embedding for each frame. 

Unlike previous work that relies only on the multi-view attention module to enhance multi-view consistency, we propose a novel \emph{3D-Aware Denoising Sampling} to further improve multi-view consistency. Specifically, we employ a feed-forward-based model conditioned on multi-view images generated by VideoMV to explicitly generate 3D models. Subsequently, these generated 3D models are rendered to the corresponding view and replace the original images produced by VideoMV in the denoising loop.



Experimental results show that VideoMV outperforms state-of-the-art multi-view synthesis approaches in terms of both efficiency and quality. For example, MVDream~\cite{Shi2023MVDreamMD} consumes 2300 GPU hours to train a 4-view generation model. In contrast, VideoMV, which uses weights from a pre-trained video generation model, only requires 4 GPU hours to train a 24-view generation model. On the other hand, VideoMV outperforms MVDream~\cite{Shi2023MVDreamMD} in metrics of image quality and multi-view consistency. 


In summary, our contributions are as follows:
\vspace{-1mm}
\begin{itemize}
\item We propose VideoMV, which is fine-tuned from off-the-shelf video generative models, for multi-view synthesis. It exhibits strong multi-view consistency behavior.
\item We introduce a novel 3D-aware denoising strategy to further improve the multi-view consistency of the generated images.
\item Extensive experiments demonstrate that our method outperforms the state-of-the-art approaches in both quantitative and qualitative results.
\end{itemize} 


\section{Related Works}

\noindent
\textbf{Distillation-based Generation.} 
Score Distillation Sampling was first proposed by DreamFusion~\cite{poole2022dreamfusion} to generate 3D models by distilling from pre-trained 2D image generative models without using any 3D data. Fantasia3D~\cite{chen2023fantasia3d} further disentangled the optimization into geometry and appearance stages. Magic3D~\cite{lin2023magic3d} uses a coarse-to-fine strategy for high-resolution 3D generation. ProlificDreamer~\cite{wang2023prolificdreamer} proposes variational score distillation (VSD), which models the 3D parameter as a random variable instead of a constant. CSD~\cite{kim2023collaborative} considers multiple samples as particles in the update and distills generative priors over a set of images synchronously. NFSD~\cite{katzir2023noisefree} proposes an interpretation that can distillate shape under a nominal CFG scale, making the generated data more realistic. SteinDreamer~\cite{wang2023steindreamer} reduces the variance in the score distillation process. LucidDreamer~\cite{liang2023luciddreamer} proposes interval score matching to counteract over-smoothing. HiFA~\cite{zhu2023hifa} and DreamTime~\cite{huang2023dreamtime} optimize the distillation formulation. RichDreamer~\cite{qiu2023richdreamer} models geometry using a multi-view normal-depth diffusion model, which makes the optimization more stable. RealFusion~\cite{melaskyriazi2023realfusion}, Make-it-3D~\cite{tang2023makeit3d}, HiFi-123~\cite{yu2023hifi123}, and Magic123~\cite{qian2023magic123} use multi-modal information to improve generation fidelity. DreamGaussian~\cite{tang2023dreamgaussian} and GaussianDreamer~\cite{yi2023gaussiandreamer} use an efficient Gaussian Splitting representation to accelerate the optimization process. However, distillation-based generation requires tens of thousands of iterations of the 2D generator and can take hours to generate a single asset.  

\noindent
\textbf{Feed-forward-based Generation.} 
Many works attempt to use a neural network to directly learn the 3D distribution by fitting 3D data. OccNet~\cite{Mescheder2018OccupancyNL} encodes shapes into function space and infers 3D structure from various inputs. MeshVAE~\cite{Tan2017VariationalAF} also learns a reasonable representation in probabilistic latent space for various applications. 3D-GAN~\cite{Wu2016LearningAP} designs a volumetric generative adversarial network for shape generation from latent space. With the development of differentiable rendering, HoloGAN~\cite{NguyenPhuoc2019HoloGANUL} and BlockGAN~\cite{NguyenPhuoc2020BlockGANL3} learn 3D representation from natural images in an unsupervised manner. To maintain multi-view consistency, some prior works~\cite{Chan2020piGANPI, Chan2021EfficientG3, Deng2021GRAMGR, Gu2021StyleNeRFAS, Niemeyer2020GIRAFFERS, Xu20213DawareIS, Zhang2022MultiViewCG} incorporate implicit 3D representations in generative adversarial networks for 3D-aware generation. GET3D~\cite{Gao2022GET3DAG}, DG3D~\cite{Zuo2023DG3DGH}, and TextField3D~\cite{Huang2023TextField3DTE} leverage DMTet~\cite{Shen2021DeepMT} for accurate textured shape modeling. Accompanied by the development of 2D diffusion models~\cite{ho2020denoising, Rombach2021HighResolutionIS}, 3D diffusion-based approaches~\cite{Liu2023MeshDiffusionSG, Kalischek2022TetraDiffusionTD, Zhou20213DSG, Luo2021DiffusionPM, Zeng2022LIONLP, Chou2022DiffusionSDFCG, Li2022DiffusionSDFTV, Cheng2022SDFusionM3, Zheng2023LocallyAS, Nam20223DLDMNI, Muller2022DiffRFR3, Gupta20233DGenTL, Shue20223DNF} use variants of diffusion models for generative shape modeling. Point-E~\cite{Nichol2022PointEAS} and Shap-E~\cite{Jun2023ShapEGC} expand the scope of the training dataset for general object generation. LRM~\cite{Hong2023LRMLR}, PF-LRM~\cite{Wang2023PFLRMPL}, and LGM~\cite{Tang2024LGMLM} choose to use a deterministic approach for reconstruction from a few views. LEAP~\cite{jiang2023leap} and FORGE~\cite{jiang2022few-forge} focus on generating the 3D model using a few images with noisy camera poses or unknown camera poses. While these approaches are many times faster than distillation-based methods, their quality is limited. 

\noindent
\textbf{Novel View Synthesis Generation.}
Some other works~\cite{sajjadi2022scene, Wiles2019SynSinEV, chan2023generative, Gu2023NerfDiffSV, Szymanowicz2023ViewsetD, Tseng2023ConsistentVS, Yu2023LongTermPC, Zhou2022SparseFusionDV, Suhail2022GeneralizablePN} combine a novel view generator with a traditional reconstruction process or a fast neural reconstruction network for 3D generation. ViewFormer~\cite{kulhánek2022viewformer} uses transformers for novel view synthesis. 3DiM~\cite{watson2022novel} is the first to use diffusion models for pose-controllable view generation. Zero123~\cite{liu2023zero1to3} adopts a large pre-trained image generator (StableDiffusion~\cite{Rombach2021HighResolutionIS}), which greatly improves generalizability after fine-tuning on Objaverse~\cite{Deitke2022ObjaverseAU}.
SyncDreamer~\cite{liu2023syncdreamer} designs a novel depth-wise attention module to generate consistent 16 views with fixed poses. Consistent123~\cite{lin2023consistent123} combines both 2D and 3D diffusion priors for 3D-consistent generation. Zero123++~\cite{Shi2023Zero123AS} overcomes common issues like texture degradation and geometric misalignment. Wonder3D~\cite{Long2023Wonder3DSI} introduces a cross-domain diffusion model. ImageDream~\cite{Wang2023ImageDreamIM} proposes global control that shapes the overall object layout and local control that fine-tunes the image details. iNVS~\cite{Kant2023RepurposingDI} enhances the novel view synthesis pipeline through accurate depth warping. MVDream~\cite{Shi2023MVDreamMD} proposes to jointly generate 4 views with dense self-attention on all views. SPAD~\cite{kant2024spad} further enhances multi-view consistency through proposed epipolar attention.

The concurrent work, IM-3D~\cite{melas20243d-im3d} and SVD~\cite{blattmann2023stable-svd}, share a similar idea of generating more consistent multi-view images. The former uses a time-consuming optimization scheme to obtain the 3D model, while the latter adopts the elevation angle instead of the complete camera pose as the condition, posing an obstacle to downstream tasks that require camera pose inputs. Compared to them, we employ a more efficient feed-forward module to obtain an explicit 3D model from noise-corrupted images. Additionally, we propose a novel \textit{3D-Aware Denoising Sampling} to further improve consistency.

\section{Method}


\begin{figure*}[t]
\begin{center}
   \includegraphics[width=1.0\linewidth]{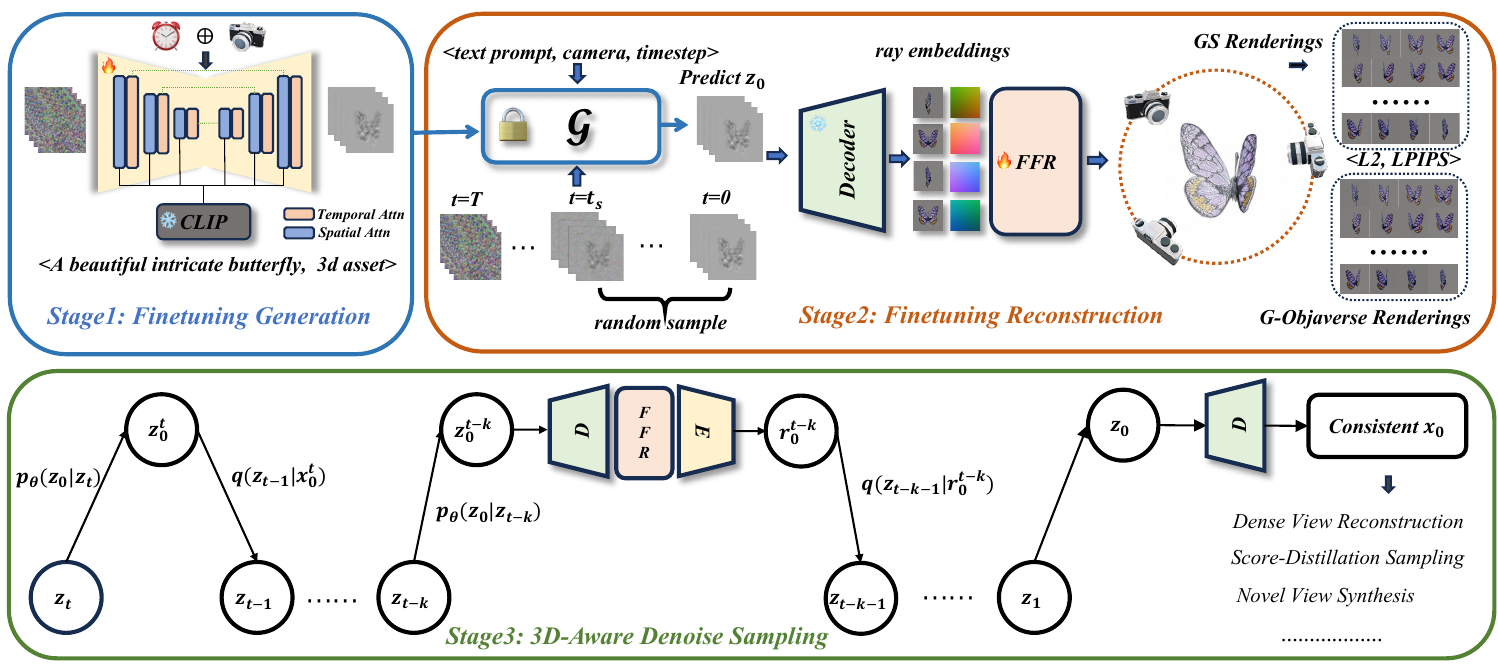}
   \caption{The overall framework. In the first stage, we take a pre-trained video generation model and fine-tune it by incorporating camera poses to generate multi-view images. Then we train a feed-forward reconstruction module to get an explicit global 3D model given noise-corrupted images. Finally, we adopt a 3D-aware denoise sampling strategy that effectively inserts the images rendered from the global 3D model into the denoising loop to further improve consistency.
   }
\label{fig:framework}
\end{center}
\vspace{-3.0em}
\end{figure*}

\subsection{Problem Statement and Approach Overview}
\label{Subsec:Overview}

\noindent
\textbf{Problem Statement.}
Given a text or single-image prompt, VideoMV aims to generate consistent multi-view images under user-specified camera poses.

\noindent
\textbf{Approach Overview.}
The key idea of VideoMV is to combine a large video generative model for initializing a multi-view generative model and a novel \textit{3D-Aware Denoising Sampling} strategy to further improve multi-view consistency. Figure~\ref{fig:framework} illustrates the pipeline of VideoMV. In the first stage, we fine-tune a pre-trained video generation model to obtain the multi-view generative model $\mathcal{G}$ (Section~\ref{sec:fine-tune}). We focus on how to specify camera poses of multi-view images to connect with object-centric videos. In the second stage, we employ a feed-forward reconstruction module to obtain an explicit global 3D model based on the images generated by $\mathcal{G}$ (Section~\ref{sec:reconstrucion}). The explicit model uses a variant of the 3D Gaussian splitting (3DGS) representation~\cite{3dgs, yi2023gaussiandreamer, Tang2024LGMLM}. In the third stage, we introduce a \textit{3D-Aware Denoising Sampling} strategy that effectively inserts the images rendered from the global 3D model into the denoising loop (Section~\ref{sec:sampling}) to further improve multi-view consistency.

\vspace{-1.0em}
\subsection{Fine-tuning Generation}
\label{sec:fine-tune}

The first stage of VideoMV fine-tunes a video generative model for multi-view image generation. This is achieved by generating videos from rendered images of 3D models for fine-tuning. In the following, we first introduce the pre-trained video generative models. We then describe how to generate video data from 3D models for fine-tuning. 

\noindent\textbf{Pre-trained video generative models.} We choose two open-source video generative models, Modelscope-T2V~\cite{wang2023modelscope} and I2VGen-XL~\cite{zhang2023i2vgen}, which are used for the text-based and single-image-based multi-view generation, respectively. Both belong to the video latent diffusion model (VLDM), which uses a pre-trained encoder and a pre-trained decoder and performs diffusion and denoising in the latent space. 

Specifically, consider a video $\bm{x} \in \mathbb{R}^{F \times H \times W \times 3}$ where $F$ is the number of frames. They use a pre-trained encoder $\mathcal{E}$ of VQGAN~\cite{vqgan} to compress it into a low-dimensional latent feature $\bm{z} = \mathcal{E}(\bm{x})$, where $\bm{z} \in \mathbb{R}^{F \times h \times w \times c}$.
In the training stage, the diffusion process samples a time step $t$ and converts $\bm{z_0}$ to $\bm{z_t}$ by injecting Gaussian noise $\epsilon$. Then a denoising network $\epsilon_{\theta}$ predicts the added noise $\epsilon_{\theta}(\bm{z_t}, y, t)$.
The corresponding optimized objective can be simplified as follows:
\begin{equation}
    \mathcal{L}_{\textup{VLDM}} = \mathbb{E}_{\bm{z_t}, y, \epsilon \in \mathcal{N}(0,1), t} \| \epsilon - \epsilon_{\theta}(\bm{z}_{t}, y, t) \|_{2}^{2},
\end{equation}
where $y$ denotes the conditional text or image. 
In the denoising sampling loop, given an initial Gaussian noise, the denoising network predicts the added noise $\epsilon_{\theta}(\bm{z_t}, y, t)$ for each step, ultimately obtaining a latent code $\bm{z_0}$, which is fed into the decoder of VQGAN~\cite{vqgan} to recover a high-fidelity video.

\noindent\textbf{Video data generation for fine-tuning.} We utilize the 3D G-Objaverse data-set~\cite{qiu2023richdreamer} to generate video data, denoted as $\bm{x}$, to fine-tune the video generation model. A key challenge is to generate data that is suitable for downstream tasks of multi-view image generation but does not present a large domain gap to the pre-trained video generation model. To this end, we generate a video of rendered images by rotating the camera around each 3D object in the G-Objaverse dataset~\cite{qiu2023richdreamer}. In our experiment, we select 24 views for each object with a fixed elevation angle (randomly selected from 5 to 30 degrees) and azimuth angles uniformly distributed between 0 and 360 degrees. 

Note that VLDM uses efficient temporal convolution and attention, which operate at the same positions between frames. This is very different from the dense attention mechanism used in MVDream~\cite{Shi2023MVDreamMD}, which operates at all positions between frames, making memory explosion for dense views generation. To utilize VLDM for fine-tuning, dense views work much better than sparse views. On the other hand, dense views offer more flexibility for downstream tasks. 

VideoMV also uses camera poses as an additional control to generate images of different viewpoints, which support arbitrary novel view synthesis.  
Inspired by previous work~\cite{Shi2023MVDreamMD, liu2023syncdreamer, Long2023Wonder3DSI}, we use a two-layer multi-layer perception (MLP) to extract a camera embedding, which is combined with the time embedding. In other words, the noise predicted by the denoising network changes to $\epsilon_{\theta}(\bm{z_t}, y, c, t)$, where $c$ denotes the camera poses. Furthermore, to maintain the generalizability of our model, we integrate additional 2D image data from LAION 2B~\cite{schuhmann2022laion}.
These images are treated as videos with the number of views set to $1$. After fine-tuning, we obtain a diffusion model, which outputs multiview images conditioned text or a single image. 

\vspace{-1.0em}
\subsection{Feed-Forward Reconstruction}
\label{sec:reconstrucion}

The second stage of VideoMV learns a neural network that reconstructs a 3D model from images generated by the model $\mathcal{G}$ trained in the first stage. In the last stage of VideoMV, we will use rendered images of this 3D model to guide the denoising step in $\mathcal{G}$ to achieve improved multi-view consistency.

We employ 3D Gaussians~\cite{3dgs} as the representation of the 3D model, which has a fast rendering pipeline for image generation. Instead of using the optimization scheme that gets 3D Gaussians parameters via fitting rendering images to input images (which is time-consuming), we employ a feed-forward manner to directly regress the attributes and number of 3D Gaussians. In the following, we first review the 3D Gaussian Splatting~\cite{3dgs} representation. We then present the reconstruction network.

\noindent\textbf{3D Gaussians.} The 3D Gaussian representation uses a set of 3D Gaussians to represent the underlying scene. Each Gaussian is parameterized by a center $\mathbf{p} \in \mathbb R^3$, a scaling factor $\mathbf{s} \in \mathbb R^3$, a rotation quaternion $\mathbf{q} \in \mathbb R^4$, an opacity value $\alpha \in \mathbb R$, and a color feature $\mathbf{c} \in \mathbb R^C$. To render the image, 3DGS projects the 3D Gaussians onto the camera imaging plane as 2D Gaussians and performs alpha compositing on each pixel in front-to-back depth order.

\noindent\textbf{Reconstruction network.} Inspired by splatter image~\cite{szymanowicz2023splatter} and LGM~\cite{Tang2024LGMLM}, we first designed a reconstruction network that learns to convert noise-corrupted multi-view latent features in the denoising procedure of $\mathcal{G}$ into Gaussian correlation feature maps, whose channel values represent the parameters of the Gaussian and whose number of pixels is equal to the 3D Gaussian number. However, we find this module difficult to learn, causing the rendered images to become blurred. One explanation is that the latent space is highly compressed, and it is difficult to learn patterns between this latent space and the underlying 3D Gaussian model. To address this issue, we adopt the decoder of VQGAN~\cite{vqgan} to decode the noise latent features into images and use these images as input for this module. For reconstruction, we employ LGM~\cite{Tang2024LGMLM} and its powerful pre-trained weights for fast training convergence. Furthermore, following LGM~\cite{Tang2024LGMLM} and DMV3D~\cite{dmv3d}, we use Plücker ray embeddings to densely encode the camera pose, and the RGB values and ray embeddings are concatenated together as input to this reconstruction module.

The task of this network is to recover global 3D even if the input multi-view images are noise-corrupted or inconsistent. Unlike LGM~\cite{Tang2024LGMLM}, which uses data augmentation strategies to simulate inconsistent artifacts of input multi-view images, we directly use the output of our multi-view generative model $\mathcal{G}$ to train the reconstruction model. In this way, we do not encounter domain gaps between the training and inference stages. 
Specifically, we train this network using the noise-corrupted images obtained by only a single denoising step of $\mathcal{G}$. The original output of $\mathcal{G}$ is the predicted noise according to the input time step $t \in [0, 1000]$, and we convert it to noise-corrupted multi-view images as training data. The details of conversion will be introduced in the next Section~\ref{sec:sampling}. In the larger timestep, the converted multi-view images are similar to Gaussian noise, which is not suitable as training data for the reconstruction network. Therefore, we select time steps in the range of $[0, t_s]$ (we set $t_s=700$) to train our module.

\vspace{-1.0em}
\subsection{3D-Aware Denoising Sampling}
\label{sec:sampling}

As shown in Figure~\ref{fig:framework}, we adopt a \textit{3D-Aware Denoising Sampling} strategy that involves the rendered images produced by our reconstruction module in a denoising loop to further improve the multi-view consistency of the resulting images. We use the DDIM~\cite{song2020denoising-ddim} scheduler with 50 denoised steps for fast sampling. The sampling step from $\bm{z_t}$ to $\bm{z_{t-1}}$ of DDIM~\cite{song2020denoising-ddim} can be formulated as follows:
\begin{equation}
    \bm{z_{t-1}} = \sqrt{\alpha_{t-1}} \underbrace{\left(\frac{\bm{z_t} - \sqrt{1 - \alpha_t} \epsilon_\theta^{(t)}(\bm{z_t)}}{\sqrt{\alpha_t}}\right)}_{\text{`` predicted }\bm{z_0} \text{''}} + \underbrace{\sqrt{1 - \alpha_{t-1} - \sigma_t^2} \cdot \epsilon_\theta^{(t)}(\bm{z_t)}}_{\text{``direction pointing to } \bm{z_t} \text{''}} + \underbrace{\sigma_t \epsilon_t}_{\text{random noise}} ,
    \label{eq:sample-eq-gen}
\end{equation}
where $\alpha_t$ and $\sigma_t$ are constants, $\epsilon_t$ is the standard Gaussian noise independent of $\bm{z_t}$, and we use $\epsilon_\theta^{(t)}$ rather than $\epsilon_{\theta}(\bm{z_t}, y, c, t)$ to denote the predicted noise for simplicity. Note that during the training of the reconstruction network, we convert the predicted noise to ``predicted $z_0$'' and decode it to $\bm{x_0}$ as the input of the training data.

\begin{table}[htb]
\scriptsize
\caption{Quantitative Comparison: Our proposal achieves consistently better performance whether in dense views (24 views) or sparse views (4 views) settings.}
\vspace{-2.0em}
\begin{center}
    \begin{tabular}{l c c c c c c}
        \toprule
         Method & PSNR$\uparrow$ & SSIM$\uparrow$ & LPIPS$\downarrow$ & ClipS & RMSE(f=1)$\downarrow$ & RMSE(f=6)$\downarrow$\\
        \hline
        \hline
        MVDream & 20.50 & 0.6708 & 0.4156 & 35.33 & 0.0637 & 0.0969\\
        VideoMV(\textbf{base}) & 22.92 & 0.7551 & 0.4107 & 35.47 & 0.0554 & 0.0963\\
        VideoMV & \textbf{23.32} & \textbf{0.7638} & \textbf{0.3682} & \textbf{35.45} & \textbf{0.
        0536} & \textbf{0.0948}\\
        \bottomrule
    \end{tabular}
\label{tab:text2mv_quantitative}
\end{center}
\vspace{-2.0em}
\end{table}

\begin{table}[htb]
\scriptsize
\caption{Quantitative comparison on image-based multi-view generation task.}
\vspace{-2.0em}

\begin{center}
    \begin{tabular}{l c c c c c c c c}
        \toprule
         Method & \multicolumn{2}{c}{PSNR$\uparrow$} & \multicolumn{2}{c}{SSIM$\uparrow$} & \multicolumn{2}{c}{LPIPS$\downarrow$} & \multicolumn{2}{c}{RMSE$\downarrow$} \\
        \hline
        
        \hline
        Zero123 & \multicolumn{2}{c}{15.36} & \multicolumn{2}{c}{0.773} & \multicolumn{2}{c}{0.1689} & \multicolumn{2}{c}{0.1404} \\
        Zero123-XL & \multicolumn{2}{c}{15.82} & \multicolumn{2}{c}{0.778} & \multicolumn{2}{c}{0.1622} & \multicolumn{2}{c}{0.1417} \\
        SyncDreamer & \multicolumn{2}{c}{16.88} & \multicolumn{2}{c}{0.790} & \multicolumn{2}{c}{0.1589} & \multicolumn{2}{c}{0.1368} \\
        VideoMV(base) & \multicolumn{2}{c}{18.06} & \multicolumn{2}{c}{0.802} & \multicolumn{2}{c}{0.1464} & \multicolumn{2}{c}{0.1326} \\
        VideoMV & \multicolumn{2}{c}{\textbf{18.24}} & \multicolumn{2}{c}{\textbf{0.809}} & \multicolumn{2}{c}{\textbf{0.1433}} & \multicolumn{2}{c}{\textbf{0.1278}} \\
        \hline
        Views & 4 & 24 & 4 & 24 & 4 & 24 & 4 & 24 \\
        \hline
        ImageDream & 11.84 & 11.41 & 0.7256 & 0.7210 & 0.3239 & 0.3367 & \textbf{0.1037} & \textbf{0.0670} \\
        AM-MV & \textbf{20.02} & \textbf{17.09} & \textbf{0.8200} & \textbf{0.7978} & \textbf{0.1382} & \textbf{0.1532} & 0.1490 & 0.0759 \\
        \bottomrule
    \end{tabular}
\label{tab:img2mv_quantitative}
\end{center}
\vspace{-2.0em}
\end{table}

\begin{figure}[htb]
  \centering
  \includegraphics[width=1.0\linewidth]{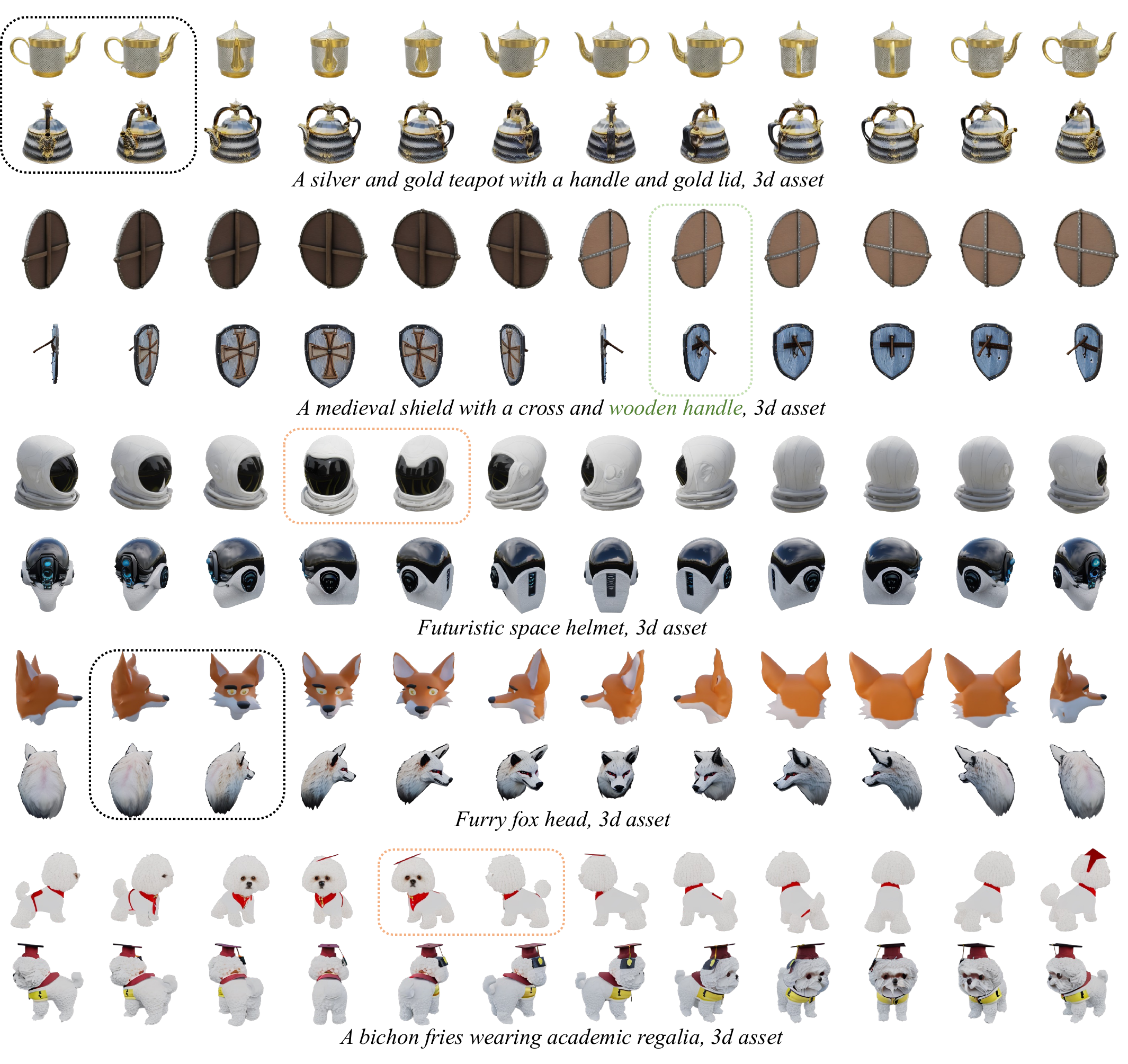}
  \vspace{-2.0em}
  \caption{Qualitative comparison of MVDream~\cite{Shi2023MVDreamMD} (\textbf{Up}) and VideoMV (\textbf{Down}) VideoMV can generate high-fidelity multi-view images which align to the text description with accurate camera control and consistent content. However, MVDream easily suffered from \textbf{inaccurate pose control} and \color{orange}{\textbf{content drifting}}. }
  \label{fig:text2mv_qualitative}
\vspace{-2.0em}
\end{figure}

In the denoising sampling loop, we employ the more consistent ``reconstructed $z_0$'' to participate in the loop, where the ``reconstructed $z_0$'' is rendered by our reconstruction module by passing ``predicted $z_0$''. 
However, this process involves decoding $\bm{z_0}$ to $\bm{x_0}$ and encoding $\bm{x_0}$ to $\bm{z_0}$, which may encounter efficiency problems. To address this issue, we use a simple strategy of using ``reconstructed $z_0$'' every $k$ timestep (we set $k=10$). We also skip it in the early denoising step. This is also reasonable since the predicted images are noising in the early steps, and thus there is no need to reconstruct.

In addition to generating multi-view images after the denoising loop, we also obtain a global 3D model represented by 3D Gaussians. 
We can convert the 3D Gaussians into a polygonal mesh, i.e., by training an efficient NeRF~\cite{NeRF, wang2021neus, wang2023neus2} from rendered images of 3D Gaussians and extracting a mesh from the density field of the resulting NeRF. 


\vspace{-1.0em}
\section{Experiments}

We perform experimental evaluation on two tasks, i.e., text-based multi-view generation and image-based multi-view generation. 
For text-based multi-view generation, we adopt MVDream~\cite{Shi2023MVDreamMD} as the baseline approach, and report metrics including PSNR, SSIM~\cite{wang2004image-ssim}, LPIPS~\cite{zhang2018unreasonable-lpips}, and flow-warping RMSE. 
For image-based multi-view generation, we adopt Zero123~\cite{liu2023zero1to3}, Zero123-XL~\cite{deitke2024objaverse-xl, liu2023zero1to3}, and SyncDreamer~\cite{liu2023syncdreamer} as baseline approaches, and report metrics that include PSNR, SSIM~\cite{wang2004image-ssim}, and LPIPS~\cite{zhang2018unreasonable-lpips}. Note that in text-based multi-view generation, we evaluate by NeRF-based novel view synthesis since no ground truth is provided.

\begin{figure}[tb]
\vspace{-1.5em}
  \centering
  \includegraphics[width=1.0\linewidth]{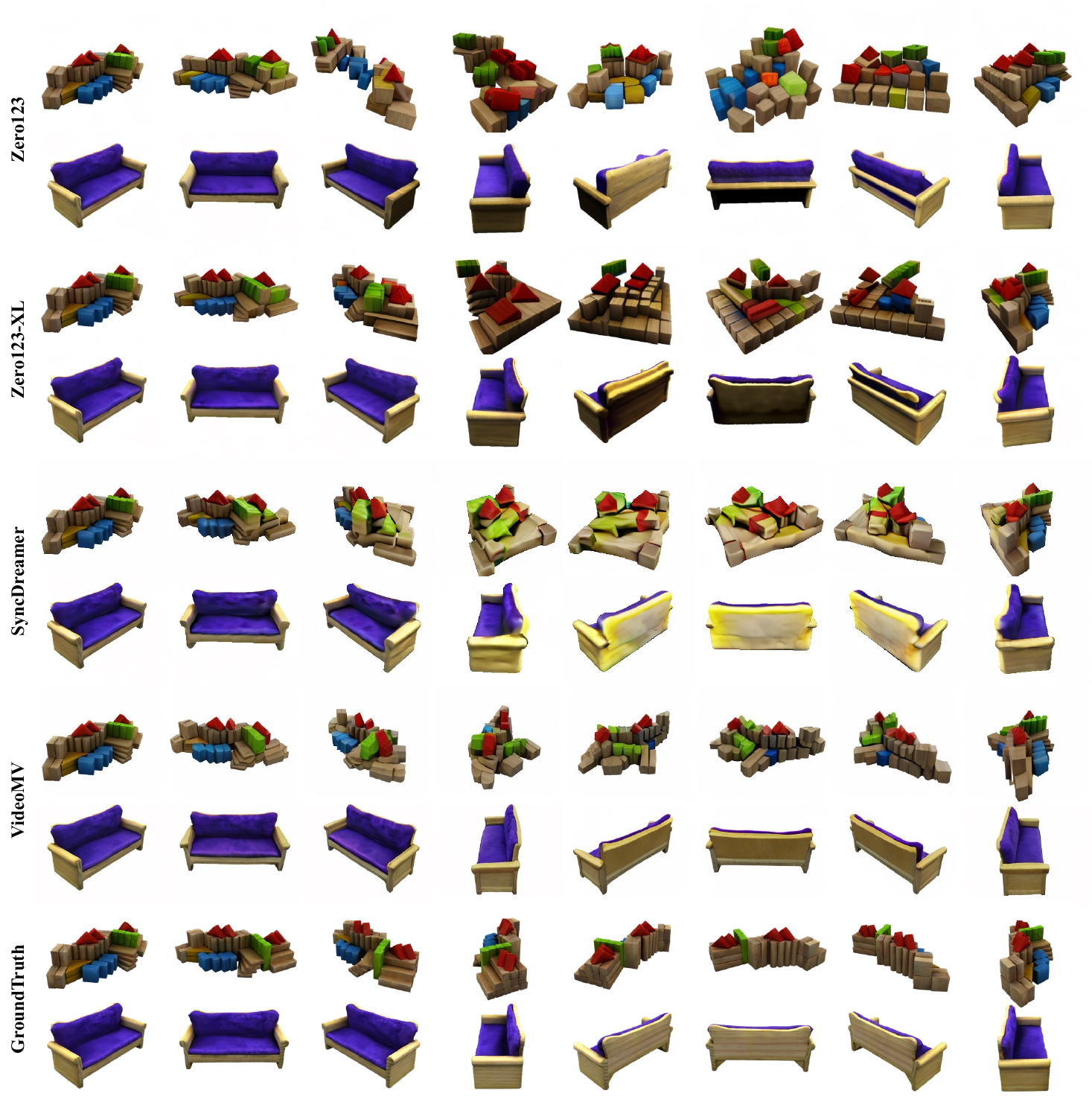}
  \caption{Qualitative comparison on GSO~\cite{gso} dataset.}
  \label{fig:img2mv_qualitative}
\vspace{-2.0em}
\end{figure}

\begin{figure}[tbp]
\vspace{-1.5em}
  \centering
  \includegraphics[width=1.0\linewidth]{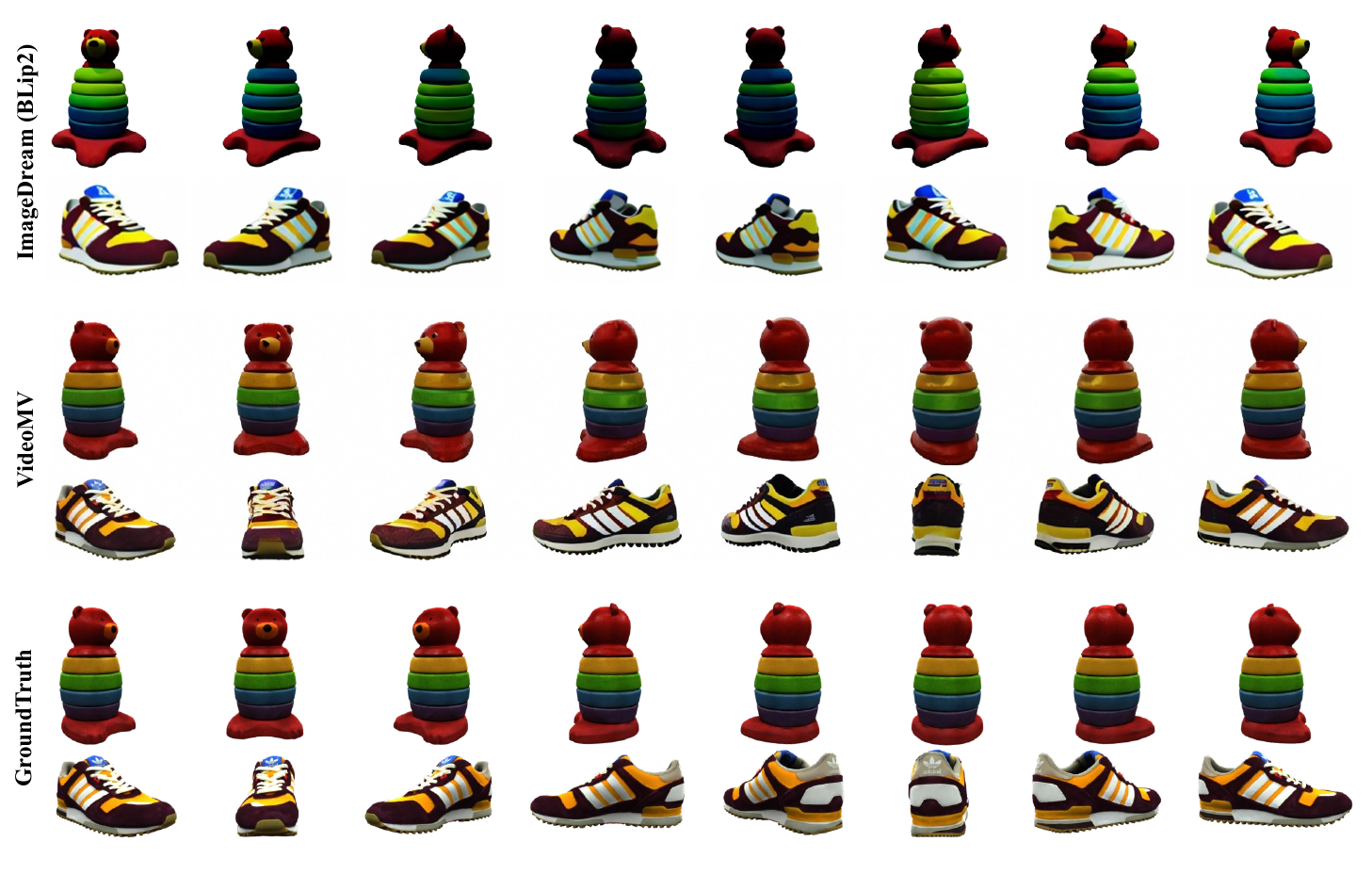}
  \caption{Qualitative comparison with ImageDream~\cite{Wang2023ImageDreamIM} on GSO~\cite{gso} dataset.}
  \label{fig:img2mv_imagedream_qualitative}
\vspace{-1.5em}
\end{figure}

\vspace{-1.0em}
\subsection{Text-based Multi-View Generation}

We use 100 single-object prompts from T3Bench~\cite{he2023t-t3} for quantitative evaluation. 
For MVDream~\cite{Shi2023MVDreamMD}, we feed circular camera poses into it and generate 24 views simultaneously. MVDream~\cite{Shi2023MVDreamMD} was trained on 32 uniformly distributed azimuth angles, and the objects were rendered twice with different random settings. Therefore, MVDream is able to generate more views interpolately given the camera poses. VideoMV was trained at 24 uniformly distributed azimuth angles, and the objects were rendered only once with random elevations. (G-Objaverse~\cite{qiu2023richdreamer}). After we generate 24 views by a specific text prompt, we use 12 views for a neural field reconstruction and report the novel view synthesis metrics (PSNR, SSIM~\cite{wang2004image-ssim}, and LPIPS~\cite{zhang2018unreasonable-lpips}) on the remaining 12 views to evaluate the multi-view consistency. We also calculated the average Clip-Score between the text prompt and generated 24 views to assess the text-to-image alignment. Another metric is flow-warping RMSE~\cite{liu2023stylerf}, which is widely adopted in 3D and video editing to evaluate semantic consistency between short-ranged or long-ranged frames. We use RAFT~\cite{teed2020raft} for optical flow estimation and softmax-splatting for warping between consecutive frames. We report Flow-Warping RMSE on two settings: one with an interval of every 1 frame and the other with an interval of every 6 frames. Note an interval of 6 frames aligns perfectly with MVDream since it is trained to produce 4 orthogonal views. We do not use 4 views for novel view synthesis evaluation since 4 views are two sparse for a reconstruction pipeline.

As depicted in \cref{tab:text2mv_quantitative}, VideoMV (\textbf{base}) significantly outperforms MVDream in 3D consistency-related metrics (PSNR, SSIM~\cite{wang2004image-ssim}, LPIPS~\cite{zhang2018unreasonable-lpips}) and flow-warping RMSE using an interval of every 1 frame. VideoMV (\textbf{base}) achieves a similar Clip-Score although trained with less data and a slightly better flow-warping RMSE using an interval of every 6 frames. By incorporating a reconstruction module for rectification, VideoMV outperforms baselines in consistency-related metrics, demonstrating the effectiveness of \textit{3D-Aware Denoising Sampling} guided by an underlying 3D model.

 Due to space constraints, we visualize some typical results with only 12 views in \cref{fig:text2mv_qualitative} for qualitative comparison with MVDream~\cite{Shi2023MVDreamMD}. We refer the readers to the supp. material for a visualization with all 24 views. Although trained with 4 views with random angles simultaneously, MVDream~\cite{Shi2023MVDreamMD} still suffered from content drifting and inaccurate pose control. In contrast, VideoMV can provide precise camera control without content drifting over dense views. VideoMV can provide consistent and fine-grained dense-view prior for downstream tasks like dense view reconstruction and distillation-based 3D generation.


\begin{figure}[tb]
\vspace{-1.0em}
  \centering
  \includegraphics[width=1.0\linewidth]{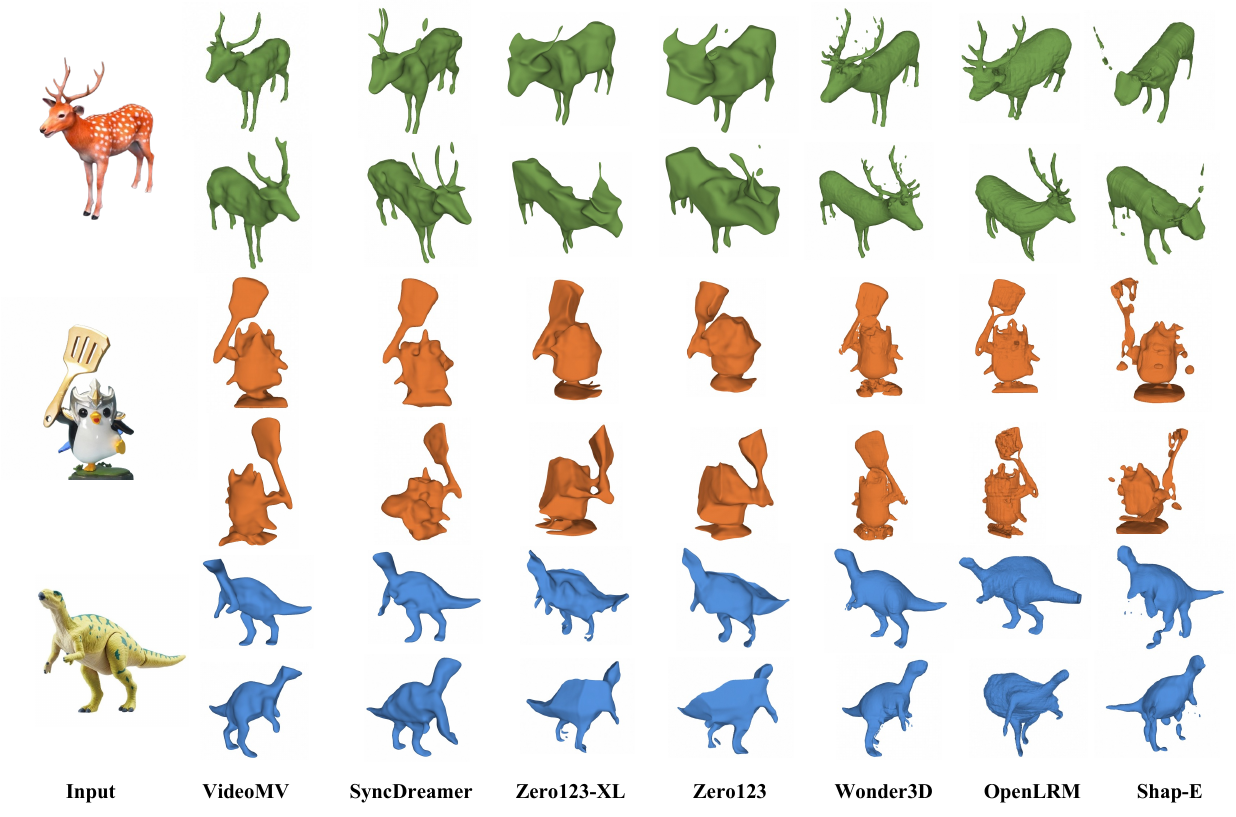}
  \caption{Reconstruction results from NVS-based methods and inference-based methods.}
  \label{fig:img2mv_mesh}
\vspace{-2.0em}
\end{figure}

\vspace{-1.0em}
\subsection{Image-based Multi-View Generation}

VideoMV can be reformulated to image-based multi-view generation. In our experiments, we fine-tune VideoMV from I2VGen-XL~\cite{zhang2023i2vgen}, which is an open-source image-based video generation method and shares the same architecture as modelscopeT2V~\cite{wang2023modelscope}. Since I2VGen-XL~\cite{zhang2023i2vgen} accepts both an input image and a text prompt, we set the text prompt to an empty string in the fine-tuning stage. We similarly train a feed-forward reconstruction module and apply consistent sampling in VideoMV. Evaluation is carried out on 50 objects from the GSO dataset (Google Scanned Objects) dataset~\cite{gso}, including the 30 objects from SyncDreamer~\cite{liu2023syncdreamer}. Since SyncDreamer~\cite{liu2023syncdreamer} is trained to generate fixed 16 views with an elevation of 30 degrees and the azimuth spans evenly in [0, 360] degrees, we only compute metrics on the $[0,3,6,9,12,15,18,21]^{th}$ frames that correspond to the $[0,2,4,6,8,10,12,14]^{th}$ frames of SyncDreamer~\cite{liu2023syncdreamer}. For Zero123~\cite{liu2023zero1to3} and Zero123-XL~\cite{deitke2024objaverse-xl}, we report metrics on the generated frames with azimuths of $[0,45,90,135,180,225,270,315]$ degrees. We also compare our method with ImageDream~\cite{Wang2023ImageDreamIM}, which is an image-prompt-based multi-view generation method. We use BLIP2~\cite{Li2023BLIP2BL} to caption the input image and evaluate under settings of 4 views and 24 views, respectively. Note that we always evaluate under the elevation settings of our baselines for fairness, which means that we use an input image of $elevation=5$ for ImageDream~\cite{Wang2023ImageDreamIM} and an input image of $elevation=30$ for SyncDreamer~\cite{liu2023syncdreamer}.

We first visualize some image-based multi-view generation results among our testing GSO dataset~\cite{gso} in \cref{fig:img2mv_qualitative}. Zero123 and Zero123-XL~\cite{liu2023zero1to3} suffer content drift since no global 3D information is utilized. SyncDreamer~\cite{liu2023syncdreamer} generates geometry-consistent multi-view images with coarse colors due to the discrete depth-wise attention applied to the low-resolution latent space. VideoMV generates more consistent results with precise colors since it adopts a global 3D representation in the full-resolution image space and utilizes the strong multi-view prior from large video generative models. The numerical results in \cref{tab:img2mv_quantitative} also consistently align with the visualization results. We find that ImageDream obtains significantly lower PSNR, SSIM, and LPIPS, but achieves better flow-warping RMSE under different settings of views. To clarify this, we also visualize the novel views generated by ImageDream in \cref{fig:img2mv_imagedream_qualitative}. As depicted, ImageDream generates novel views based on the input image and text prompt, which produces prompt-aligned multi-view images but does not consistently follow the pixel-level constraint of the input image. Moreover, it also suffers from inaccurate pose control and content drifting problems since it is based on MVDream~\cite{Shi2023MVDreamMD}. It achieves better flow-warping RMSE in the 24 views setting, since it sometimes produces consecutive images with the same pose(see the samples of ImageDream in \cref{fig:img2mv_imagedream_qualitative}). Despite these shortcomings, ImageDream maintains better semantic consistency under the 4 views setting, which makes it more suitable for distillation sampling than VideoMV. 



\vspace{-1.0em}
\subsection{Analysis on 3D-Aware Denoising Sampling}

\begin{figure}[htb]
  \centering
  \vspace{-1.0em}
  \includegraphics[width=1.0\linewidth]{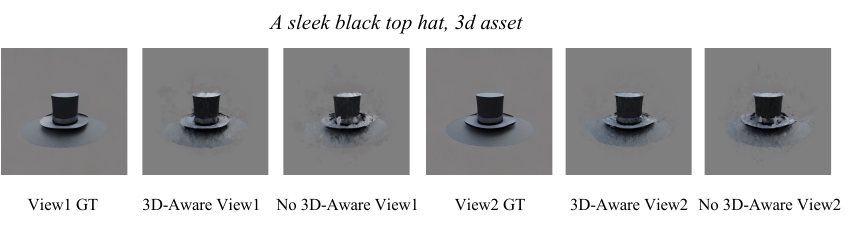}
  \vspace{-2.0em}
  \caption{Ablation for \textit{3D-Aware Denoising Sampling}.}
  \label{fig:ablation}
  \vspace{-2.0em}
\end{figure}

We have shown in \cref{tab:text2mv_quantitative} and \cref{tab:img2mv_quantitative} that our proposed \textit{3D-Aware Denoising Sampling} faithfully improves the consistency-related metrics. However, observing the changes in images before and after applying \textit{3D-Aware Denoising Sampling} does not reveal a clear increase of consistency in the human vision system (we will provide the difference maps in the supp. material). Alternatively, we visualize the novel view synthesis results, which use 12 views for training and another 12 views for evaluation. As shown in \cref{fig:ablation}, \textit{3D-Aware Denoising Sampling} (3D-Aware) significantly improves the perceptual quality of novel views over baseline (No 3D-Aware). This shows the effectiveness of our proposed \textit{3D-Aware Denoising Sampling}.
\begin{figure}[htb]
  \centering
  \includegraphics[width=1.0\linewidth]{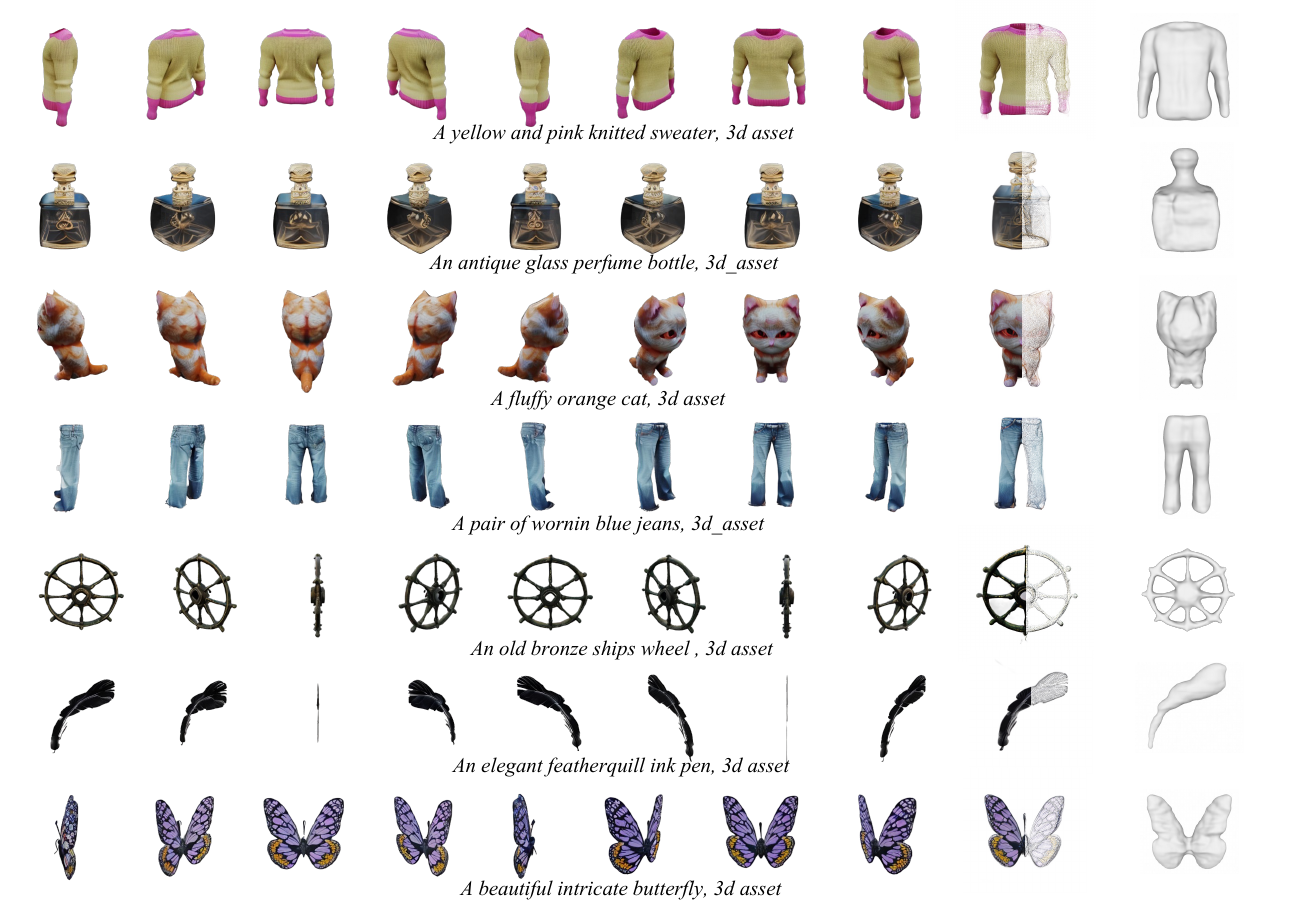}
  \caption{Text-based reconstruction results of VideoMV.}
  \label{fig:app_text2mv}
\vspace{-2.0em}
\end{figure}

\section{Applications}

\begin{figure}[htb]
  \centering
  \includegraphics[width=1.0\linewidth]{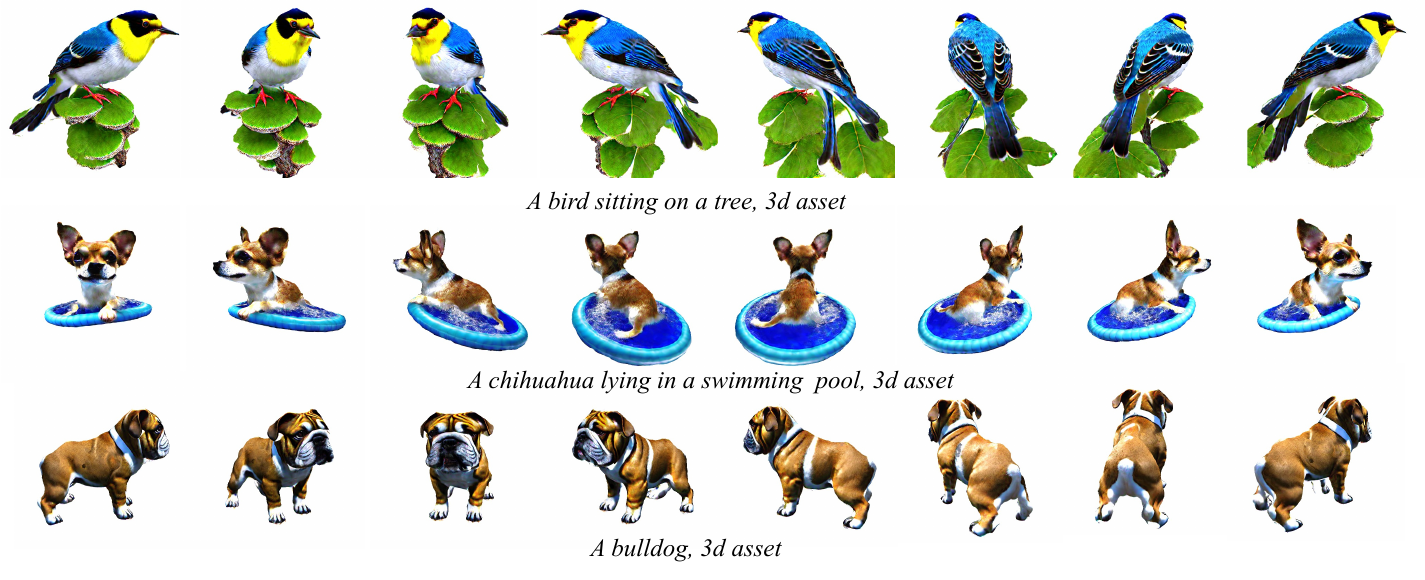}
  \vspace{-2.0em}
  \caption{Visualization of multi-view score distillation.}
  \label{fig:app_sds}
\end{figure}

\vspace{-1.0em}
\subsection{Dense View Reconstruction}
Our proposal can generate 24 dense views with specified camera poses using DDIM~\cite{song2020denoising-ddim} sampling in 5 seconds. The dense view generated is enough for a reconstruction pipeline such as NeRF~\cite{NeRF} or Neus~\cite{wang2021neus}. We show reconstruction results on both image-based multi-view generation and text-based multi-view generation tasks.

For the image-based multiview generation task, we visualize results from NVS-based methods(Zero123(XL)~\cite{liu2023zero1to3}, SyncDreamer~\cite{liu2023syncdreamer}) and reconstruction-focused methods(Wonder3D~\cite{Long2023Wonder3DSI}, OpenLRM~\cite{Hong2023LRMLR, openlrm}, and Shap-E~\cite{Jun2023ShapEGC}) accompanied with our approach. As shown in \cref{fig:img2mv_mesh}, VideoMV achieves better reconstruction results among NVS-based methods due to the highly consistent dense view produced by our pipeline. Wonder3D~\cite{Long2023Wonder3DSI} produces shapes with more details for the use of predicted normal maps, but sometimes produce floating artifacts. Inference-based methods such as Shap-E and OpenLRM may produce shapes that are not well aligned with the input images. The multi-view images generated by VideoMV can provide competitive prior for 3D generation compared with image-based NVS methods. Although the Neus~\cite{wang2021neus} reconstruction is smooth and does not provide geometry details, we can adopt the cross-domain attention mechanism proposed in Wonder3D~\cite{Long2023Wonder3DSI} to produce aligned normal maps and enhance the performance of our dense view reconstruction.

For the text-based multi-view generation task, we visualize the results of VideoMV only. As depicted in \cref{fig:app_text2mv}, VideoMV can also recover geometry from multi-view images generated from text prompts by Neus~\cite{wang2021neus}. As a by-product, we can also produce a Gaussian splatting field~\cite{Kerbl20233DGS} from multi-view images in seconds. 

\vspace{-1.0em}
\subsection{Distillation-based Generation}
Our proposal can also be applied as the priori of score distillation sampling~\cite{poole2022dreamfusion}. As shown in \cref{fig:app_sds}, we can distillate faithful shapes and texture from a multi-view score distillation loss and avoid the Janus problem most of the time. Note that we are focusing on consistent multi-view image generation, so we do not fully optimize the distillation pipeline. Distillation from dense views is also an interesting task for future work.

\vspace{-1.0em}
\section{Conclusions}
\vspace{-1.0em}
In this paper, we present a consistent dense multi-view generation method that can generate 24 views at various elevation angles. By fine-tuning large video generative models for several GPU hours, our proposal can effectively produce dense and consistent multi-view images from an input image or a text prompt. Future directions may focus on developing a robust neural reconstruction pipeline based on the provided consistent dense views. Moreover, we have shown that there are rich opportunities in connecting videos and multi-view based 3D vision tasks. We hope our findings in turning a video generative model into a consistent multi-view image generator can also inspire other 3D generation and video-related tasks. 

%
%
\bibliographystyle{splncs04}
\bibliography{main}

\newpage



\centerline{\Large Supplementary Material}

\section{Architecture Details}

We provide comprehensive information on the architectures of the networks employed. In~\cref{arch:modelscope}, we elaborate on the text-based video generation network (ModelScopeT2V~\cite{wang2023modelscope}), including its transformation into a text-based multi-view image generator. In~\cref{arch:i2vgen}, we discuss the image-based video generation network (I2VGen-XL~\cite{zhang2023i2vgen}) and its conversion into an image-based multi-view image generator. Finally, in~\cref{arch:LGM}, we present our large GaussianSplatting-based reconstruction model (LGM~\cite{Tang2024LGMLM}) and how it is utilized for noise reconstruction fine-tuning.

\begin{figure}[htbp]
\vspace{-1.0em}
  \centering
  \includegraphics[width=1.0\linewidth]{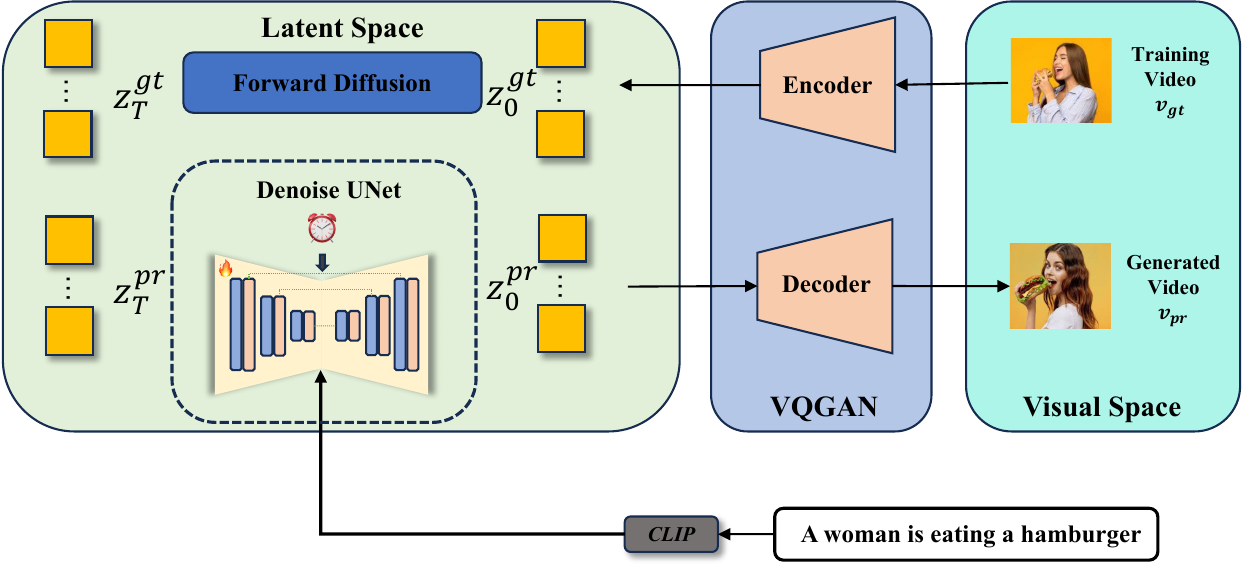}
  \caption{Overview of the architecture of ModelScopeT2V}
  \label{fig:arch_modelscopet2v}
\vspace{-2.0em}
\end{figure}

\subsection{ModelScopeT2V~(Text-Based Video Generation)}
\label{arch:modelscope}
The main paradigm of ModelScopeT2V is shown in \cref{fig:arch_modelscopet2v}. Overall, it comprises two main components: VQGAN and Denoise UNet. VQGAN aims to reconstruct the original video as precisely as possible. Given a training video $v_{gt}$, the encoder of VQGAN compresses it into the latent space as $z_{0}^{gt}$. Then, a quantizer is applied to $z_{0}^{gt}$. Note that the function of the quantizer is to find the closest quantized discrete vector of $z_{0}^{gt}$, and we denote the output from the quantizer as $q_{0}^{gt}$. Finally, the decoder of VQGAN takes $q_{0}^{gt}$ as input and outputs the reconstruction $v_{0}^{rec}$. The overall objective of VQGAN is the weighted summation of the quantization loss between $L_{q} = f_{q}(z_{0}^{gt}, q_{0}^{gt})$, the reconstruction loss $L_{r} = f_{r}(v_{0}^{gt}, v_{0}^{rec})$, and an optional adversarial loss $L_{adv} = f_{adv}(v_{0}^{gt}, v_{0}^{rec})$.

\begin{figure}[htbp]
\vspace{-1.0em}
  \centering
  \includegraphics[width=1.0\linewidth]{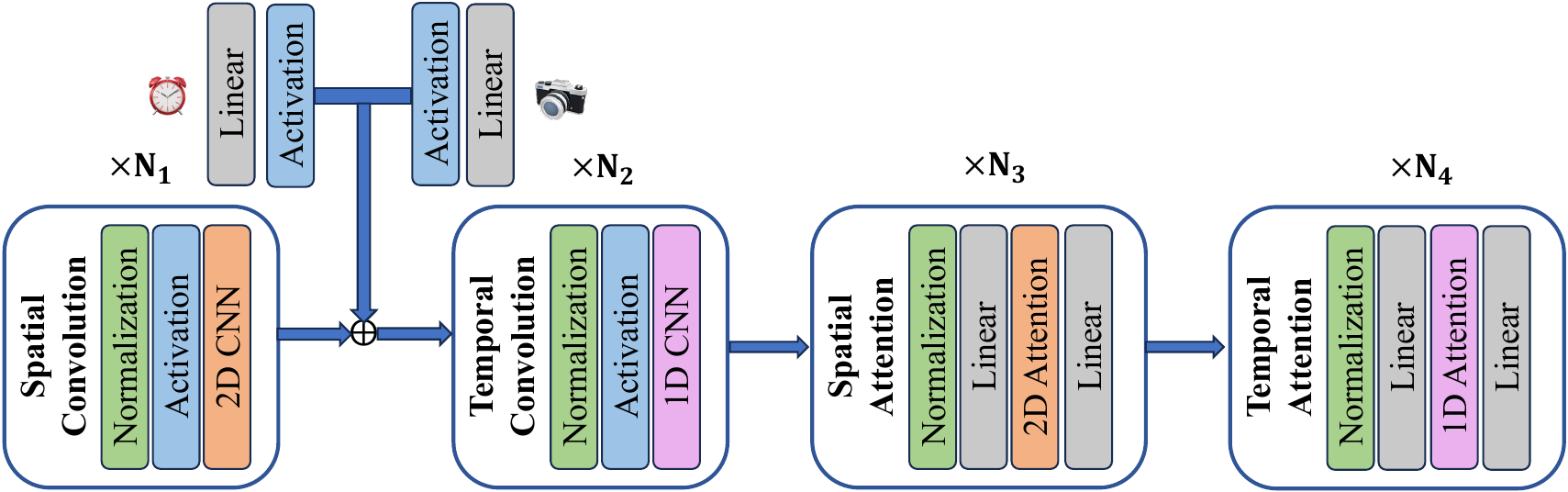}
  \caption{Buiding blocks of the Denoise UNet.}
  \label{fig:blocks_modelscopet2v}
\vspace{-2.0em}
\end{figure}

DenoiseUNet aims to recover $z_{0}^{gt}$ from the noise-corrupted $z_{T}^{gt}$. The optimization objective is the denoise reconstruction loss $L_{dr} = f_{dr}(z_{0}^{gt}, z_{0}^{pr})$. $z_{0}^{pr}$ is predicted by the network mapping function $\epsilon_{\theta}$ as $z_{0}^{pr} = \epsilon_{\theta}(z_{t}^{gt}, y, t)$. As depicted in \cref{fig:blocks_modelscopet2v}, DenoiseUNet is built with several Spatial-Temporal convolutional and attention modules. In the original implementation of ModelScopeT2V~\cite{wang2023modelscope}, timesteps $t$ are injected into the spatial convolution modules as residuals. To convert ModelScopeT2V~\cite{wang2023modelscope} into a text-based multi-view generator, we pass camera poses through an activation layer and a linear layer. The output of these layers has the same feature dimension as the embeddings of timesteps, and we further add them together. Afterward, the added embeddings of timesteps and camera poses are projected to specific feature dimensions and added together with the output of the Spatial Convolution module. Another modification is the input data. The original input of ModelScopeT2V accepts a video with 32 frames, while we modify the frame number to 24 for multi-view image generation.

\begin{figure}[htbp]
\vspace{-1.0em}
  \centering
  \includegraphics[width=1.0\linewidth]{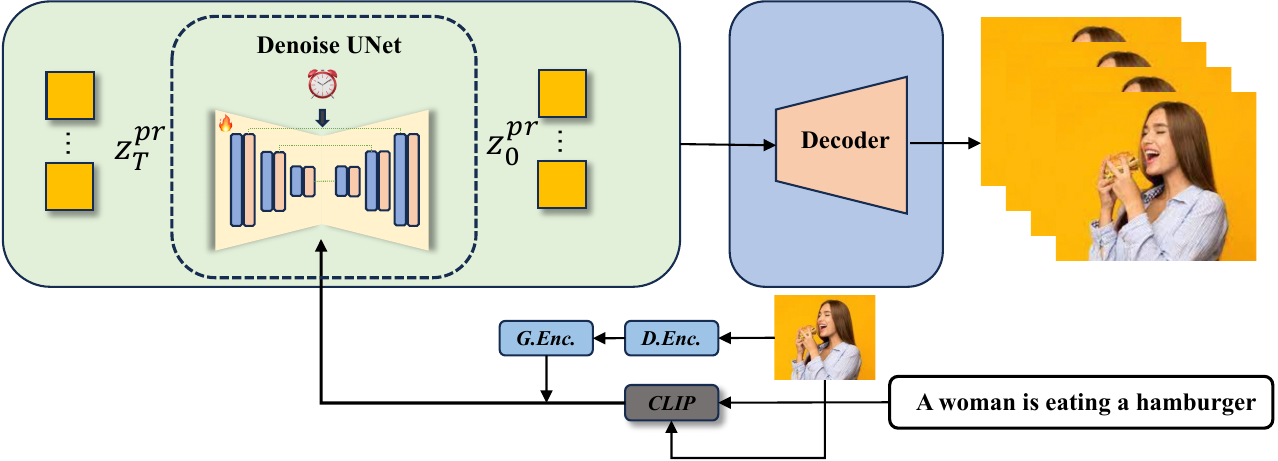}
  \caption{Overview of the architecture of I2VGen-XL.}
  \label{fig:arch_i2v}
\vspace{-2.0em}
\end{figure}
 
\subsection{I2VGen-XL~(Image-Based Video Generation)}
\label{arch:i2vgen}

The primary paradigm of I2VGen-XL is illustrated in~\cref{fig:arch_i2v}. It is important to note that the framework follows a one-staged approach, distinguishing it from the two-staged architecture proposed by \cite{zhang2023i2vgen}. Upon careful examination, we discovered that the available open-source implementation actually corresponds to a one-staged model; hence, we decided to adopt this version as depicted in \cref{fig:arch_i2v}. This one-staged model allows for the incorporation of both images and text as conditions, providing additional global (\textbf{$G.Enc.$}) and detailed (\textbf{$D.Enc$}) information extracted from the image.

I2VGen-XL~\cite{zhang2023i2vgen} shares the same DenoiseUNet architecture and VQGAN architecture with ModelScopeT2V~\cite{wang2023modelscope}. However, they differ in terms of dataset utilization and condition injection methodology. To convert I2VGen-XL~\cite{zhang2023i2vgen} into a multi-view image generator, we designated the input prompt as an empty string. For instance, we replaced 'A woman is eating a hamburger' with an empty prompt. Furthermore, we adjusted the frame number to 24 and set the training video resolution to 256$\times$256 when fine-tuning a multi-view image generator based on I2VGen-XL~\cite{zhang2023i2vgen}. We observed that augmenting the training dataset size yields improvements in terms of generalizability. Specifically, our implemented image-based multi-view generation model was trained on a dataset comprising approximately 170K samples, carefully curated by excluding textureless instances from G-Objaverse. It is important to note that this dataset differs from the high-quality 28K text-video pairs utilized for fine-tuning our text-to-video model.

\begin{figure}[htbp]
\vspace{-1.0em}
  \centering
  \includegraphics[width=0.8\linewidth]{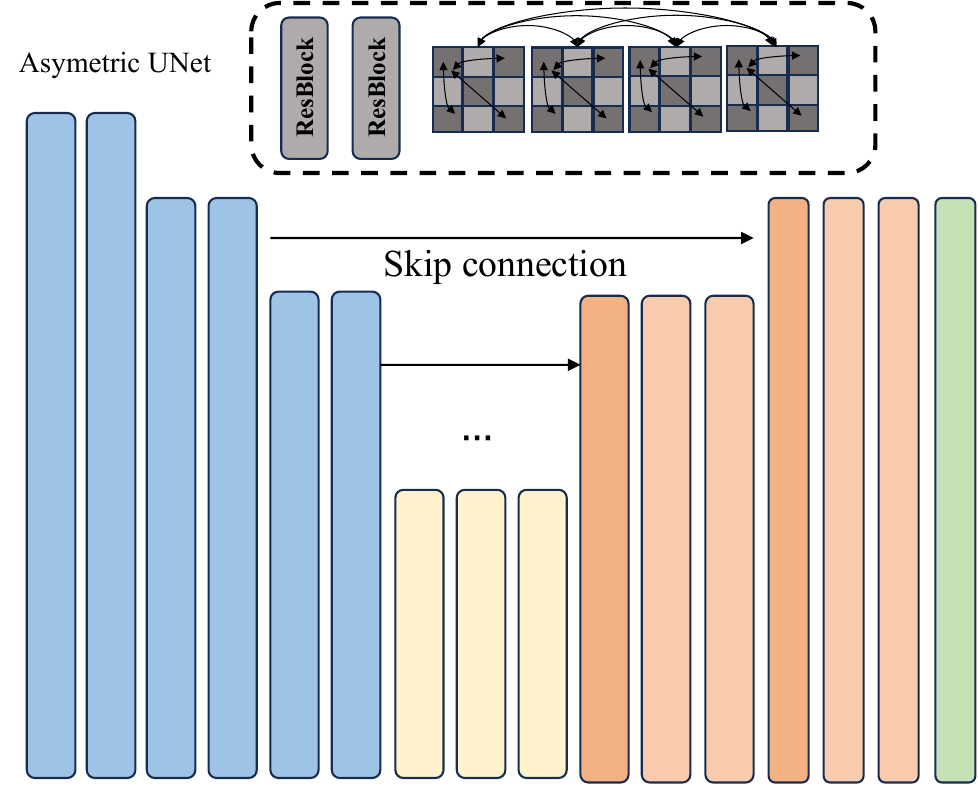}
  \caption{Architecture of the Asymmetric UNet.}
  \label{fig:lgm}
\vspace{-2.0em}
\end{figure}

\subsection{Feed-Forward Reconstruction}
\label{arch:LGM}
We have adopted the identical architecture proposed in LGM~\cite{Tang2024LGMLM}. As illustrated in~\cref{fig:lgm}, this asymmetric UNet architecture offers advantages in terms of memory efficiency by mitigating the increase in points within the GaussianSplatting representation caused by high-resolution output. It incorporates dense self-attention, similar to MVDream~\cite{Shi2023MVDreamMD}.
Considering computing resources, we did not extend it to accommodate 24 views. Yet, we have future plans for developing a dense view reconstruction model.

Sequentially, we fine-tune the asymmetric UNet using the "predicted $x_0$". Inspired by LGM~\cite{Tang2024LGMLM}, we randomly select four views from the output of VideoMV as inputs for FFR, resulting in a Gaussian field. One notable distinction is that we incorporate supervision by rendering all 24 views, deviating from the original implementation which utilized only 8 views. Additionally, we use a background color of $[128,128,128]$ instead of $[255,255,255]$ for Text-to-Multi-view image generation. On the other hand, for Image-to-Multi-view image generation, we still model the background as pure white. During the inference stage, we use fixed 4 orthogonal views for stable reconstruction performance, similar to LGM~\cite{Tang2024LGMLM}. For further details on the configuration of this sparse view reconstruction pipeline, we suggest referring to the original implementation~\cite{Tang2024LGMLM}.
\begin{figure}[htbp]
\vspace{-1.0em}
  \centering
  \includegraphics[width=1.0\linewidth]{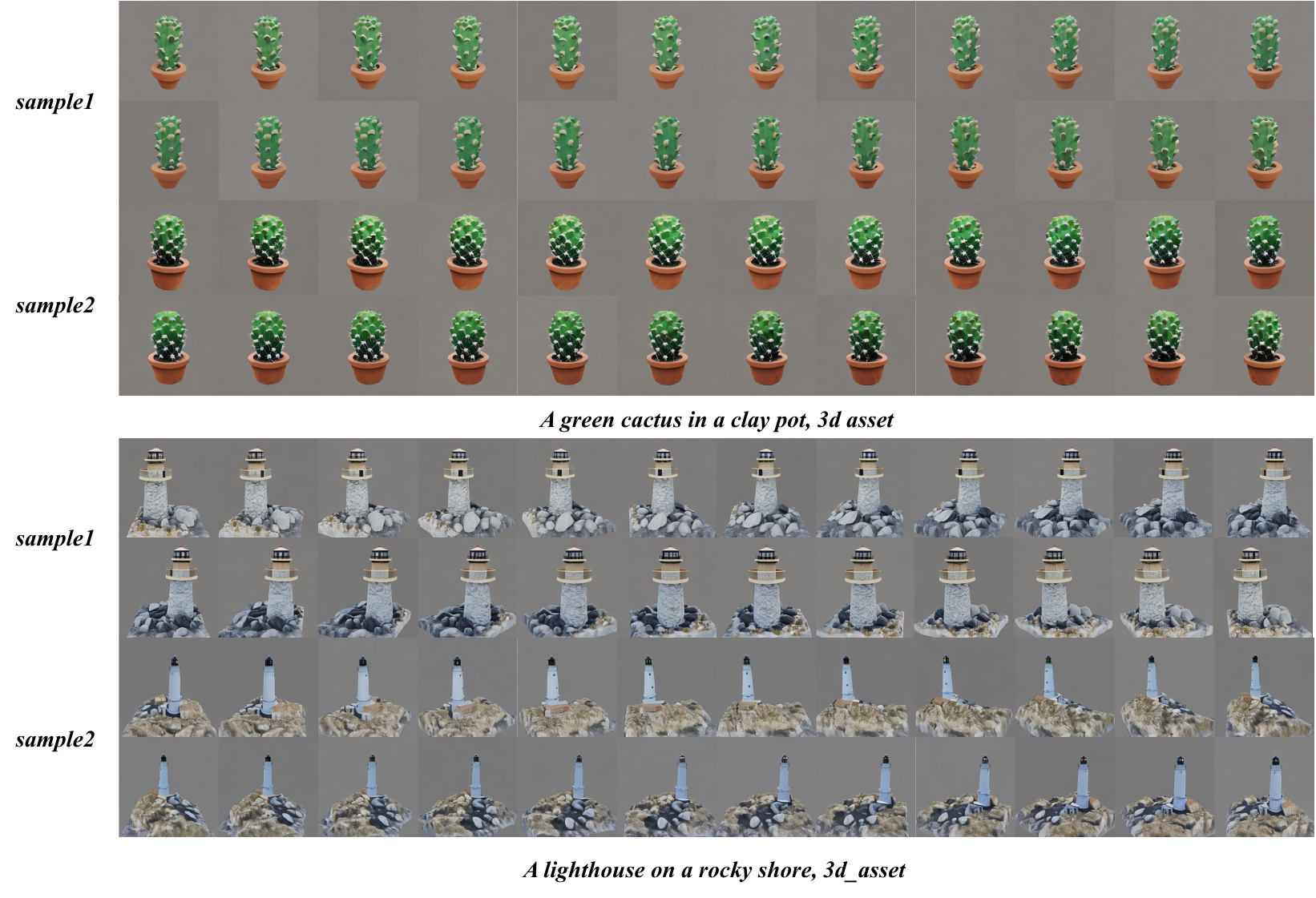}
  \caption{Visual various results of Text-to-Multi-View image generation on T3Bench~(Part I).}
  \label{fig:supp_various1}
\vspace{-2.0em}
\end{figure}

\section{Experiment Details}
In our implementation of VideoMV, we maintain a learning rate of $3e^{-5}$ and a batch size of 32. We utilize the AdamW optimizer and employ FP16 for efficient gradient descent without weight decay. The training dataset comprises 28K samples for text-conditioned VideoMV and 170K samples for image-conditioned VideoMV, all sourced from G-Objaverse~\cite{qiu2023richdreamer}. For text-conditioned VideoMV, the training process converges within half an hour using 8 NVIDIA A100 GPUs. With further training, the performance slightly improves. Image-based VideoMV requires a total training time of 24 hours utilizing 8 NVIDIA A100 GPUs.

\section{Text-to-Multi-View Image Generation}

\begin{figure}[htbp]
  \centering
  \includegraphics[width=1.0\linewidth]{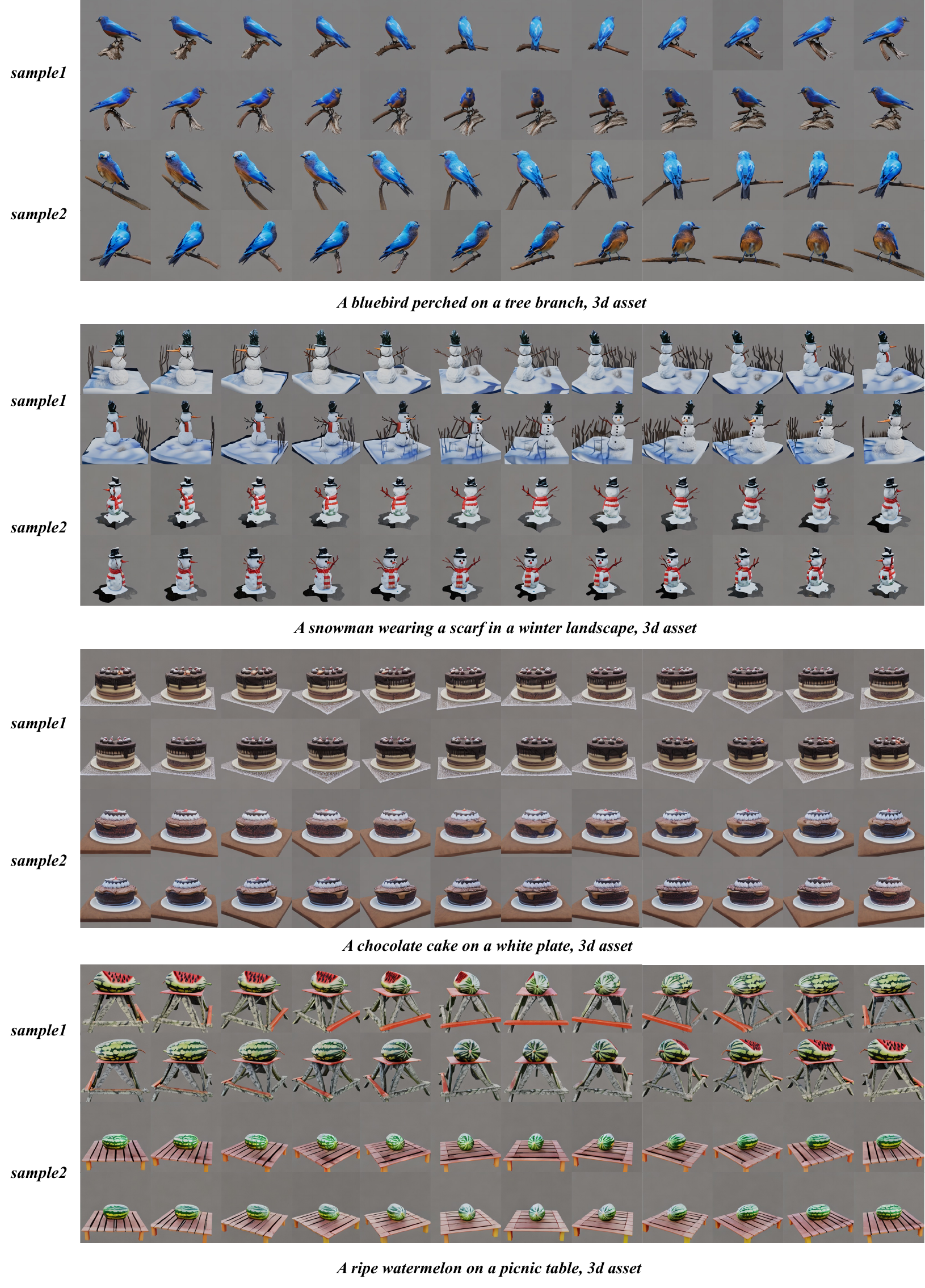}
  \caption{Visual various results of Text-to-Multi-View image generation on T3Bench~(Part II).}
  \label{fig:supp_various2}
\end{figure}

\begin{figure}[htbp]
  \centering
  \includegraphics[width=1.0\linewidth]{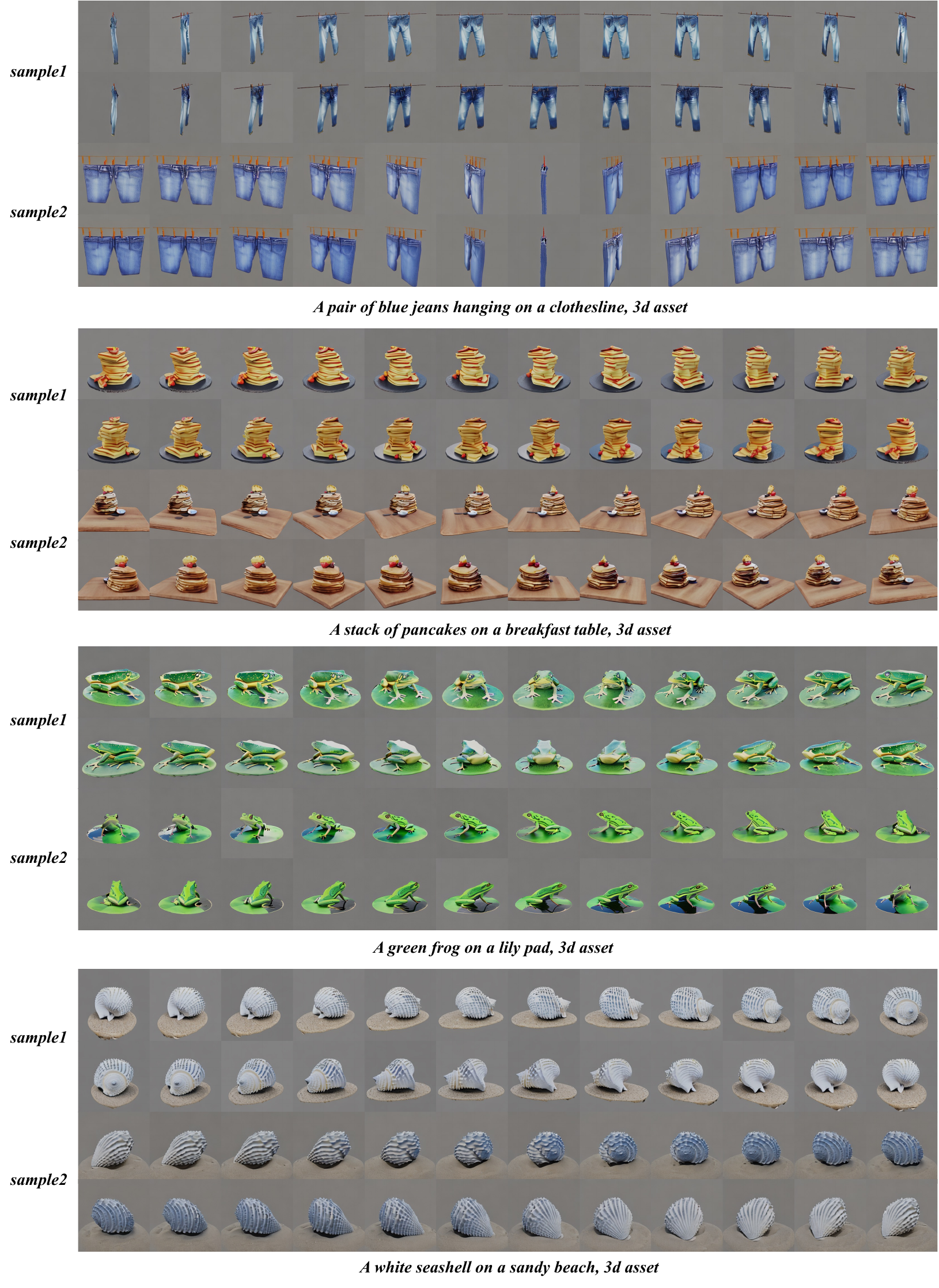}
  \caption{Visual various results of Text-to-Multi-View image generation on T3Bench~(Part III).}
  \label{fig:supp_various3}
\end{figure}

\begin{figure}[htbp]
  \centering
  \includegraphics[width=1.0\linewidth]{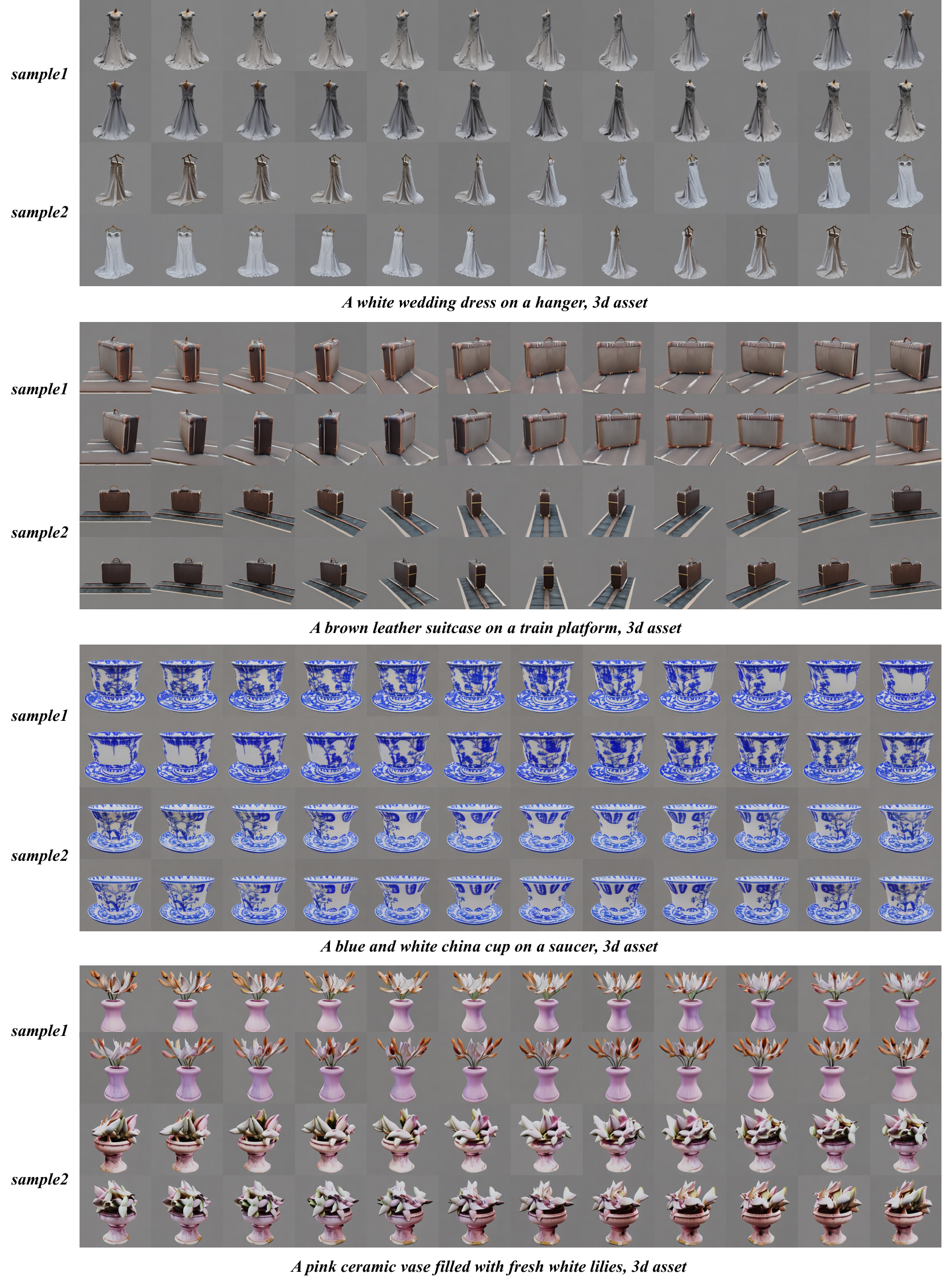}
  \caption{Visual various results of Text-to-Multi-View image generation on T3Bench~(Part IV).}
  \label{fig:supp_various4}
\end{figure}

\begin{figure}[htbp]
  \centering
  \includegraphics[width=1.0\linewidth]{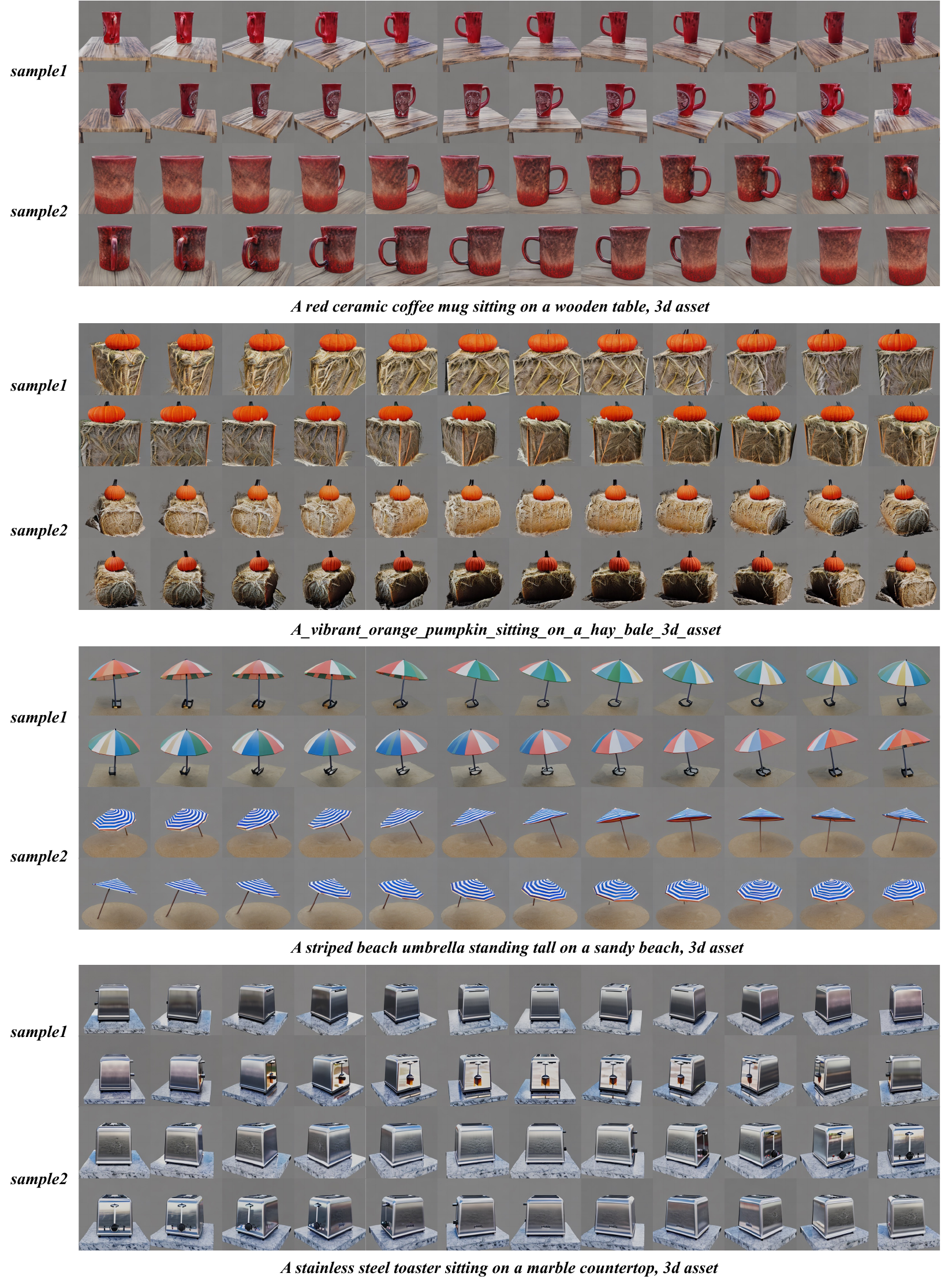}
  \caption{Visual various results of Text-to-Multi-View image generation on T3Bench.~(Part V)}
  \label{fig:supp_various5}
\end{figure}

The VideoMV demonstrates the ability to generate diverse outcomes by employing different random noises while maintaining the same prompt. As depicted from \cref{fig:supp_various1} to \cref{fig:supp_various5}, VideoMV produces a range of astonishing results across various prompts selected from the multi-object list in T3Bench~\cite{he2023t-t3}.

\subsection{More Qualitative Results}
\begin{figure}[htbp]
  \centering
  \includegraphics[width=1.0\linewidth]{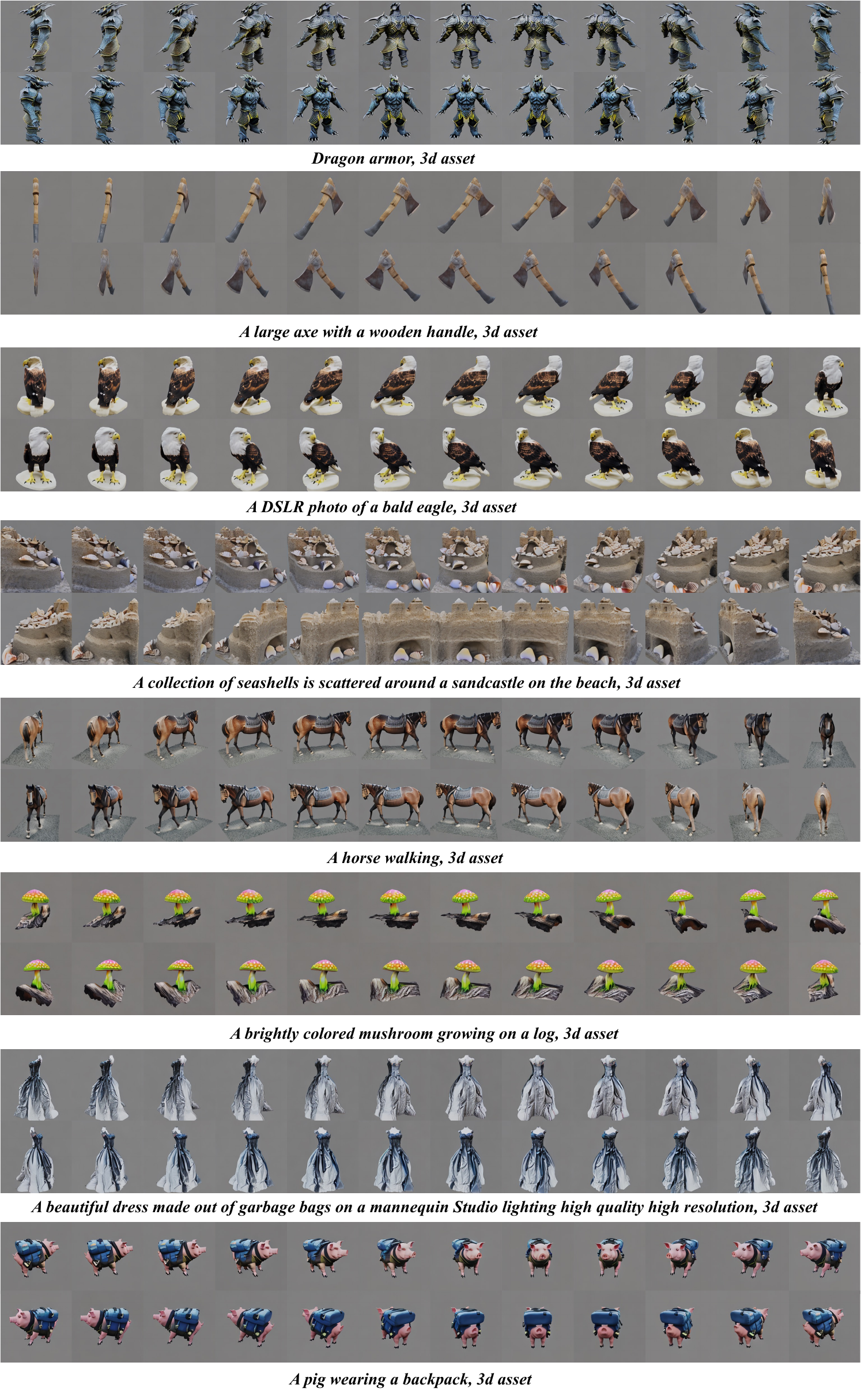}
  \vspace{-8mm}
  \caption{Visual results of Text-to-Multi-View image generation~(Part I), prompts from DreamFusion.}
  \label{fig:qua1}
\end{figure}

\begin{figure}[htbp]
  \centering
  \includegraphics[width=1.0\linewidth]{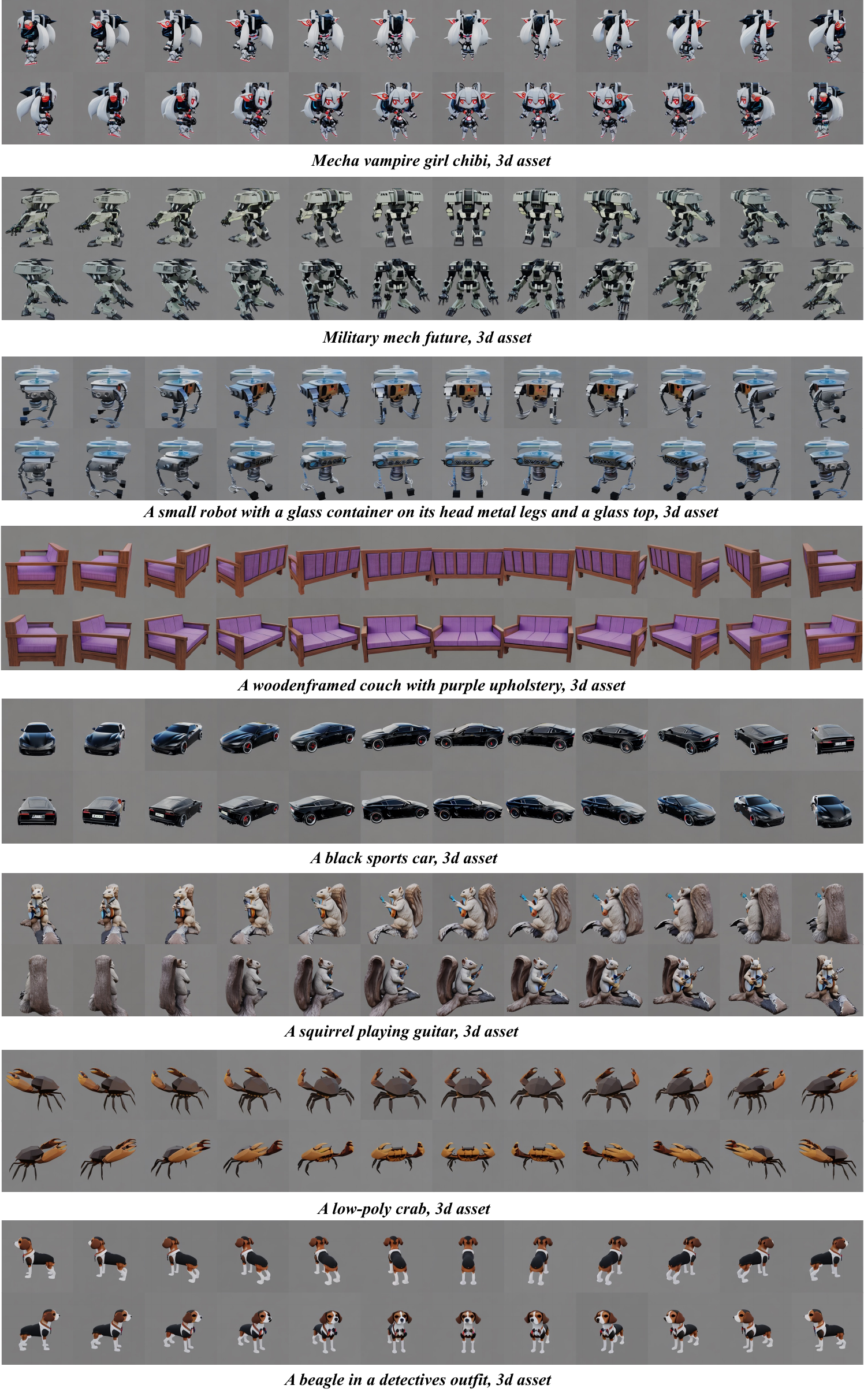}
  \vspace{-8mm}
  \caption{Visual results of Text-to-Multi-View image generation~(Part II), prompts from DreamFusion.}
  \label{fig:qua2}
\end{figure}

\begin{figure}[htbp]
  \centering
  \includegraphics[width=1.0\linewidth]{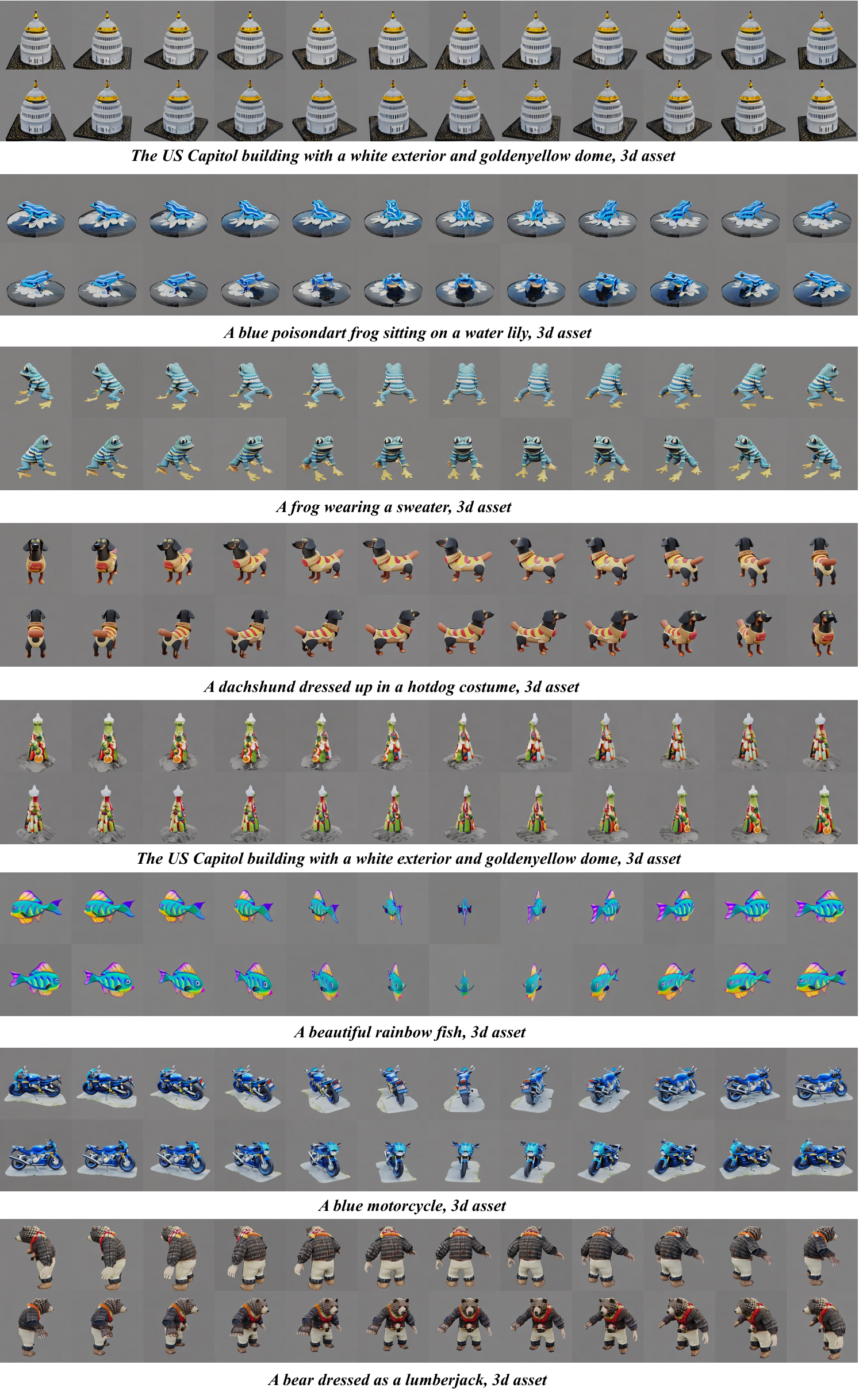}
  \vspace{-8mm}
  \caption{Visual results of Text-to-Multi-View image generation~(Part III), prompts from DreamFusion.}
  \label{fig:qua3}
\end{figure}

\begin{figure}[htbp]
  \centering
  \includegraphics[width=1.0\linewidth]{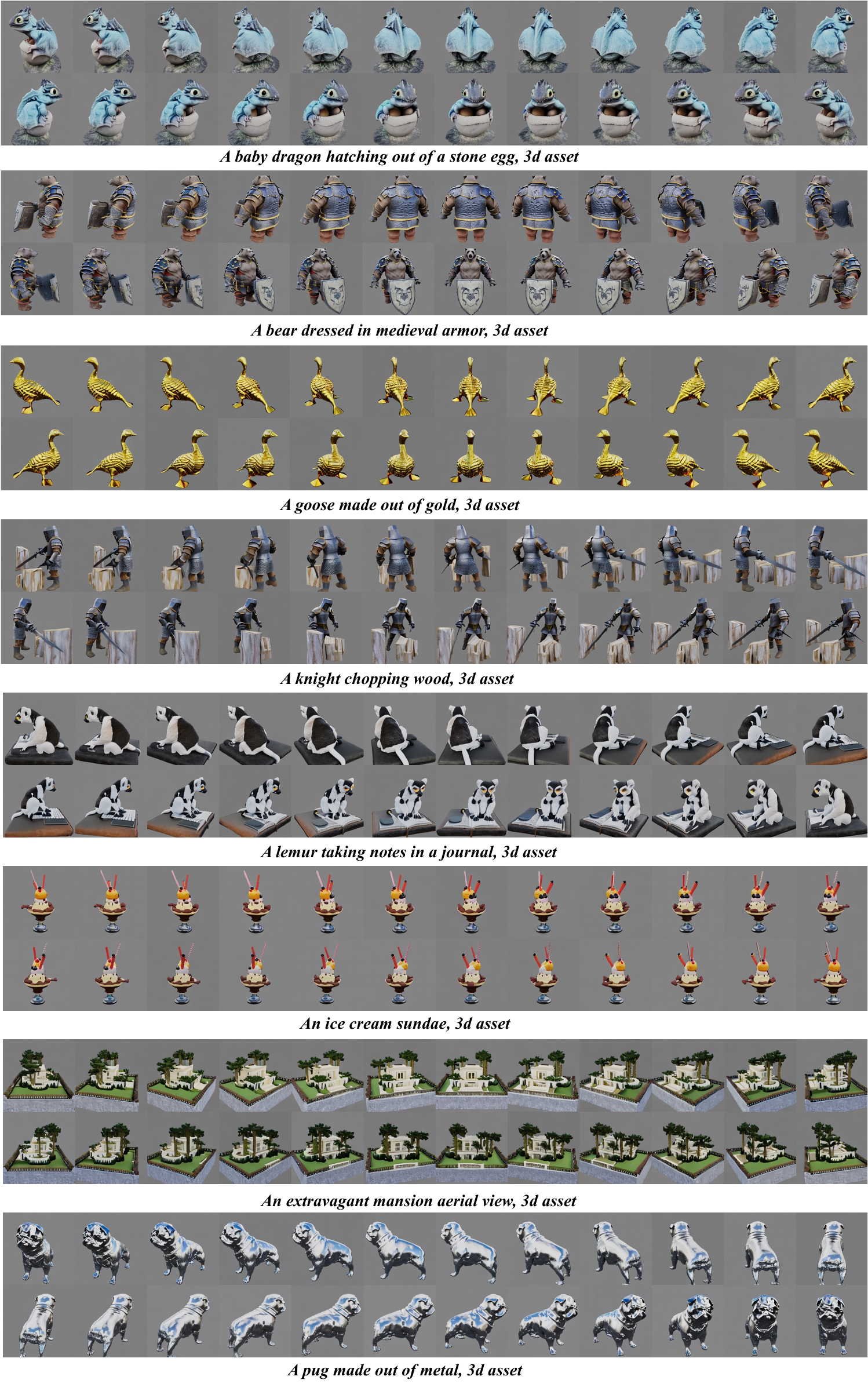}
  \vspace{-5mm}
  \caption{Visual results of Text-to-Multi-View image generation~(Part IV), prompts from DreamFusion.}
  \label{fig:qua4}
\end{figure}

\begin{figure}[htbp]
  \centering
  \includegraphics[width=1.0\linewidth]{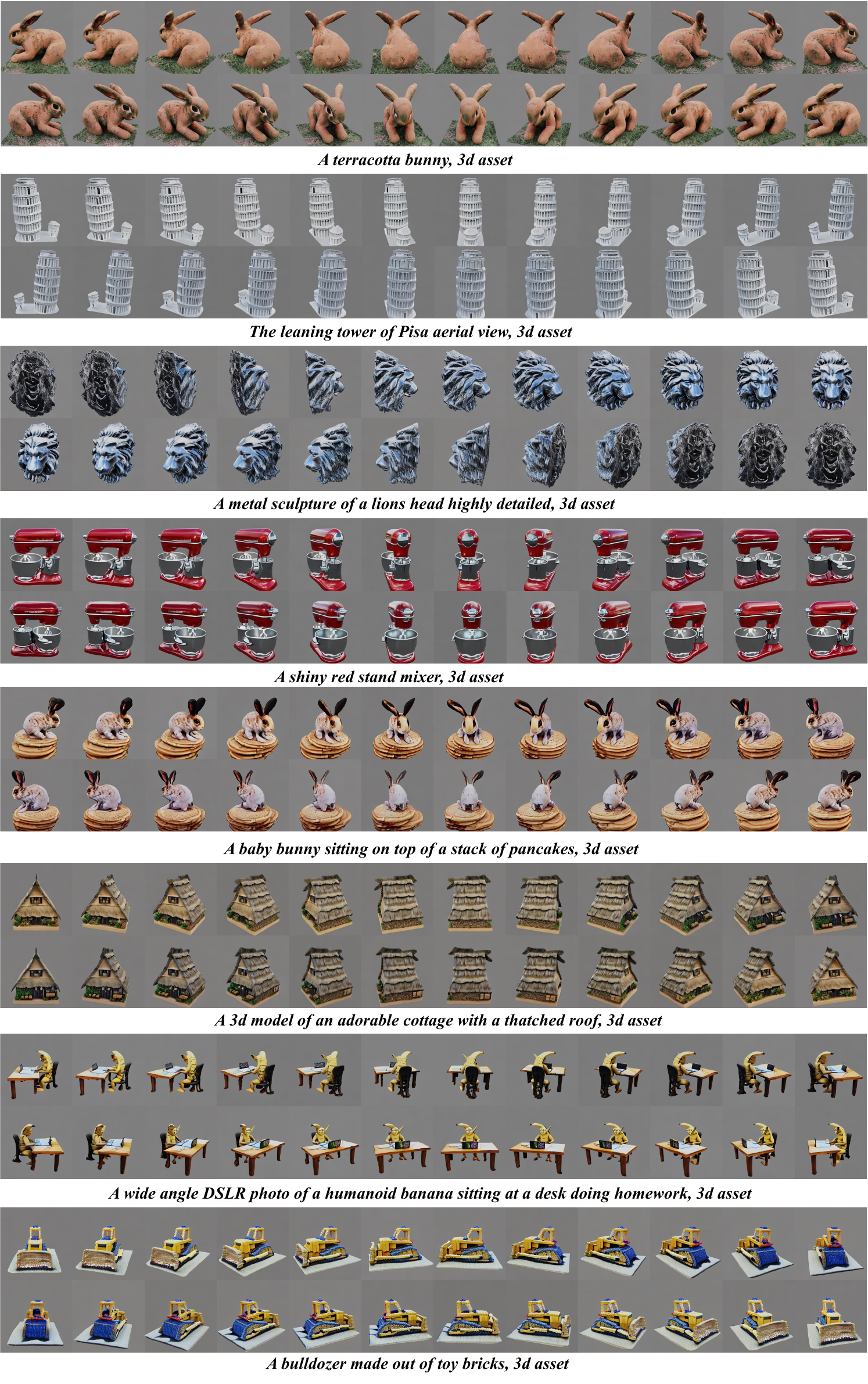}
  \caption{Visual results of Text-to-Multi-View image generation~(Part V), prompts from DreamFusion.}
  \label{fig:qua5}
\end{figure}

\begin{figure}[htbp]
  \centering
  \includegraphics[width=1.0\linewidth]{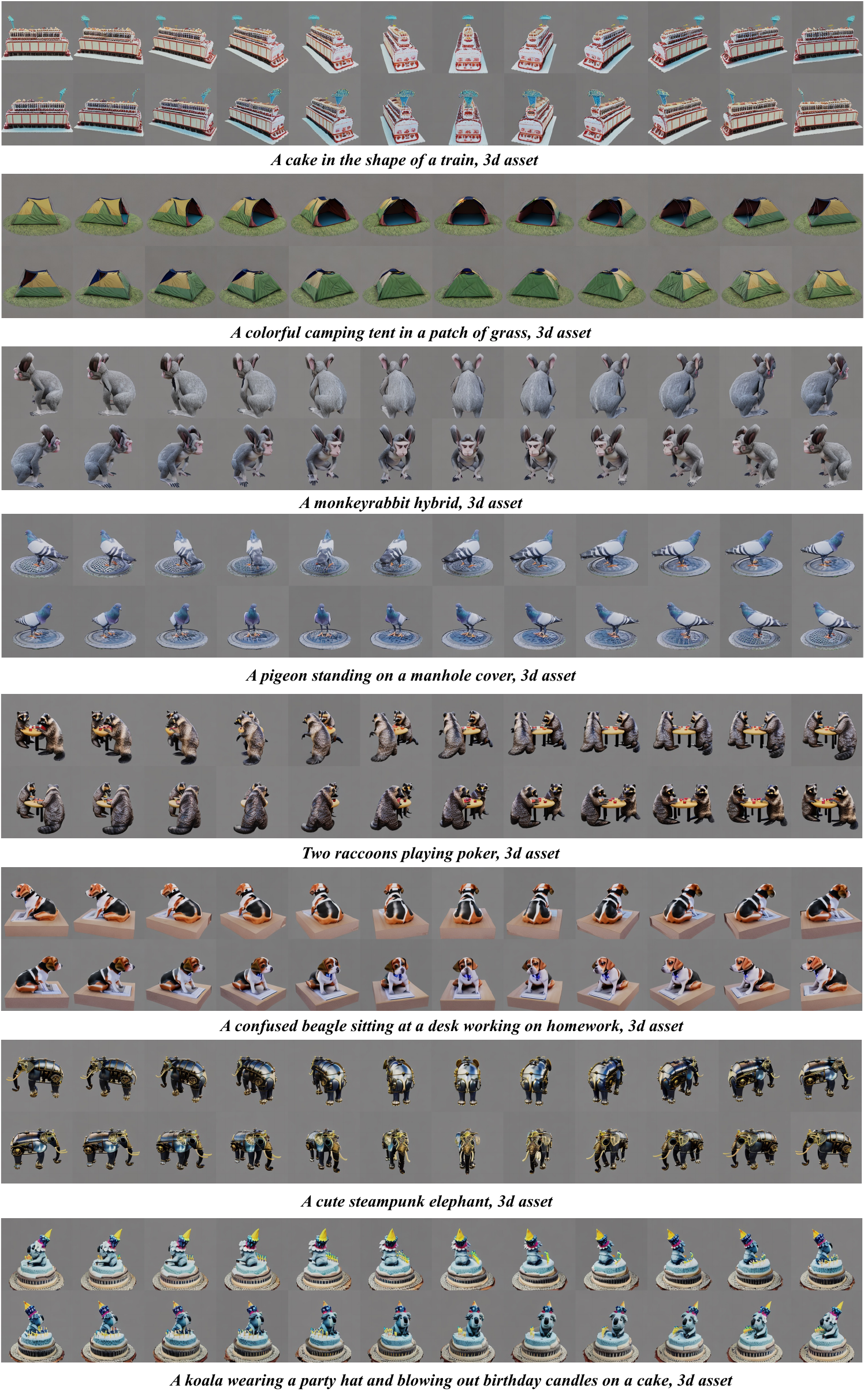}
  \vspace{-8mm}
  \caption{Visual results of Text-to-Multi-View image generation~(Part VI), prompts from DreamFusion.}
  \label{fig:qua6}
\end{figure}

\begin{figure}[htbp]
  \centering
  \includegraphics[width=1.0\linewidth]{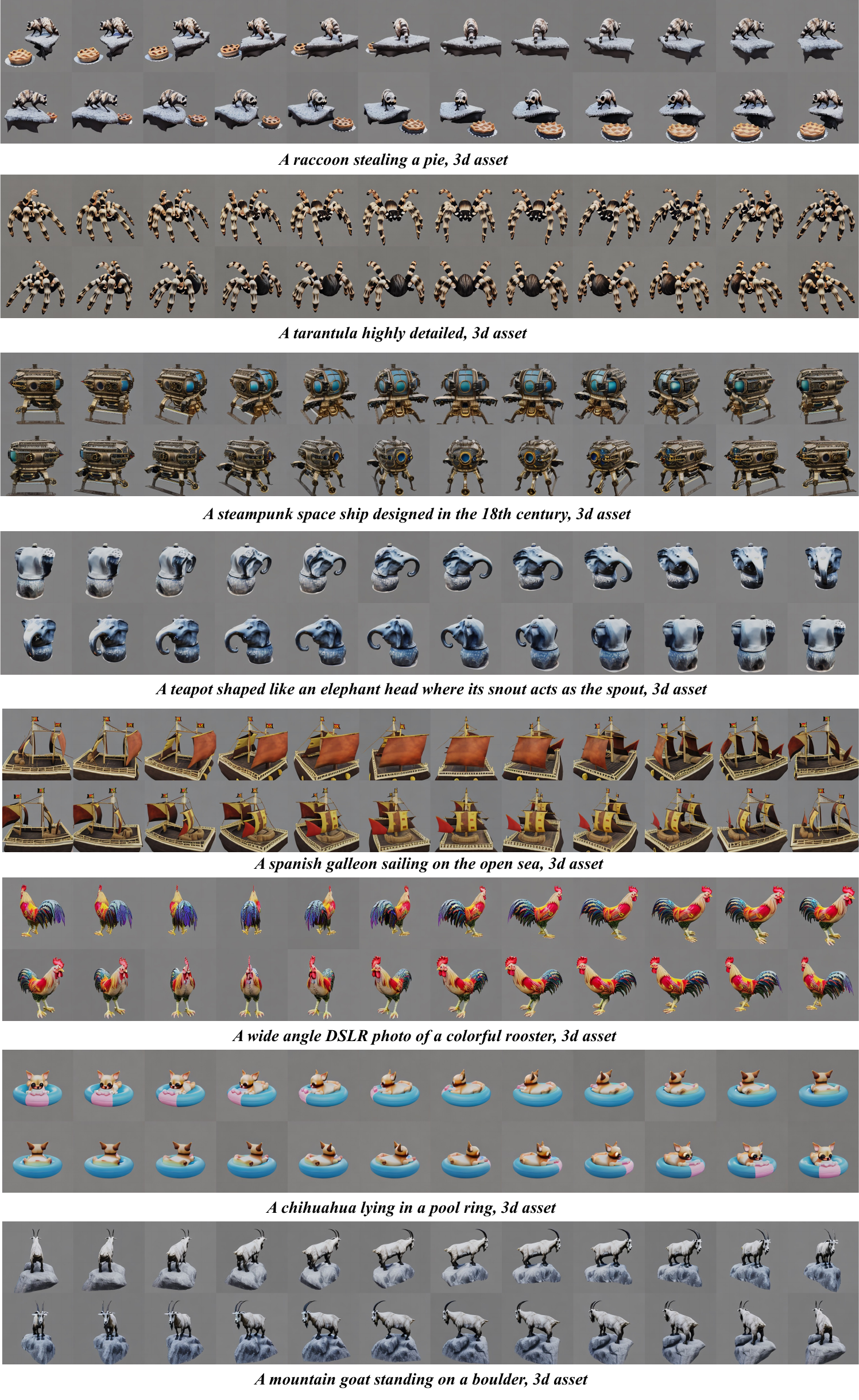}
  \vspace{-8mm}
  \caption{Visual results of Text-to-Multi-View image generation~(Part VII), prompts from DreamFusion.}
  \label{fig:qua7}
\end{figure}

\begin{figure}[htbp]
  \centering
  \includegraphics[width=1.0\linewidth]{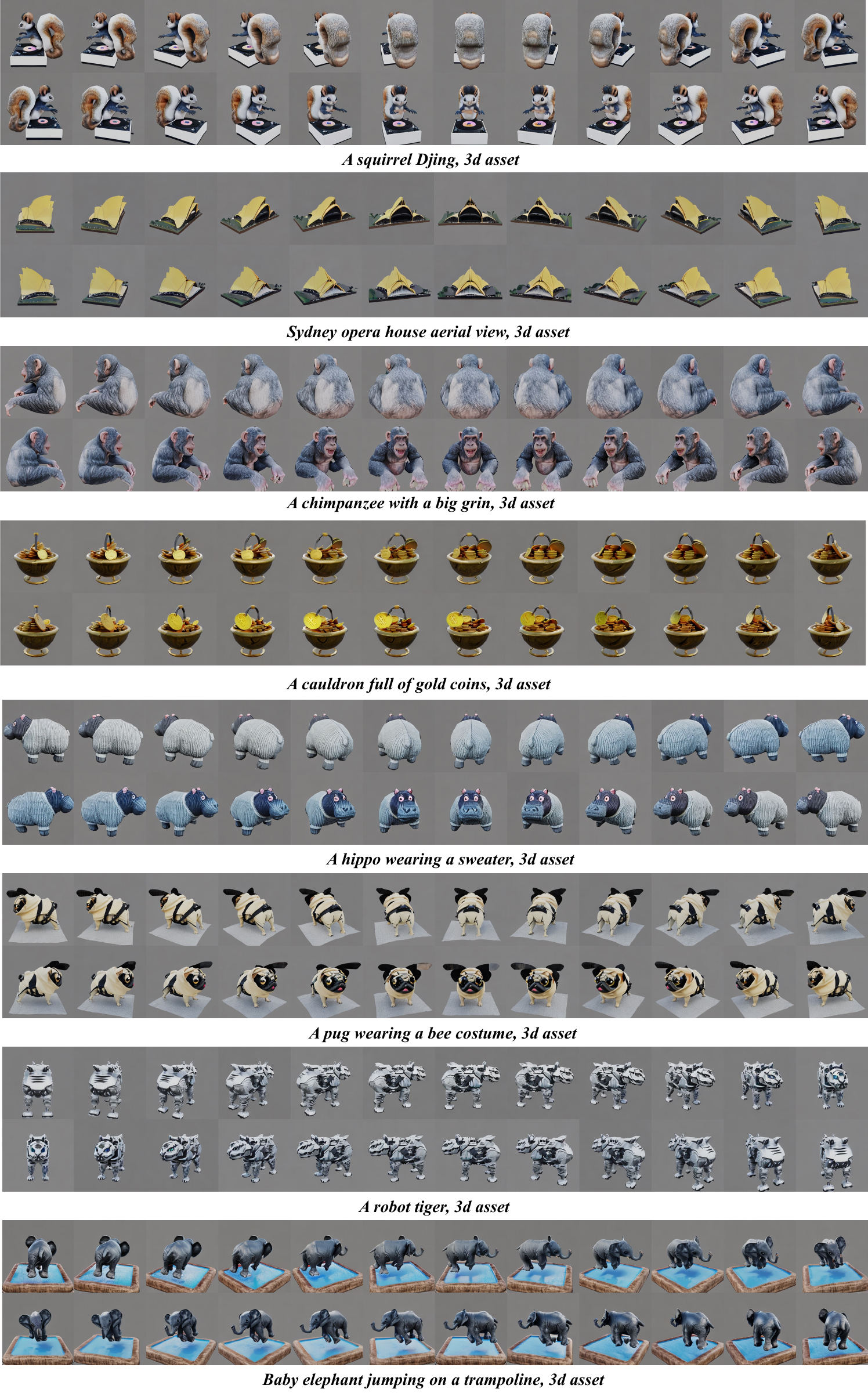}
  \vspace{-5mm}
  \caption{Visual results of Text-to-Multi-View image generation~(Part VIII), prompts from DreamFusion.}
  \label{fig:qua8}
\end{figure}

\begin{figure}[htbp]
\vspace{-1.0em}
  \centering
  \includegraphics[width=1.0\linewidth]{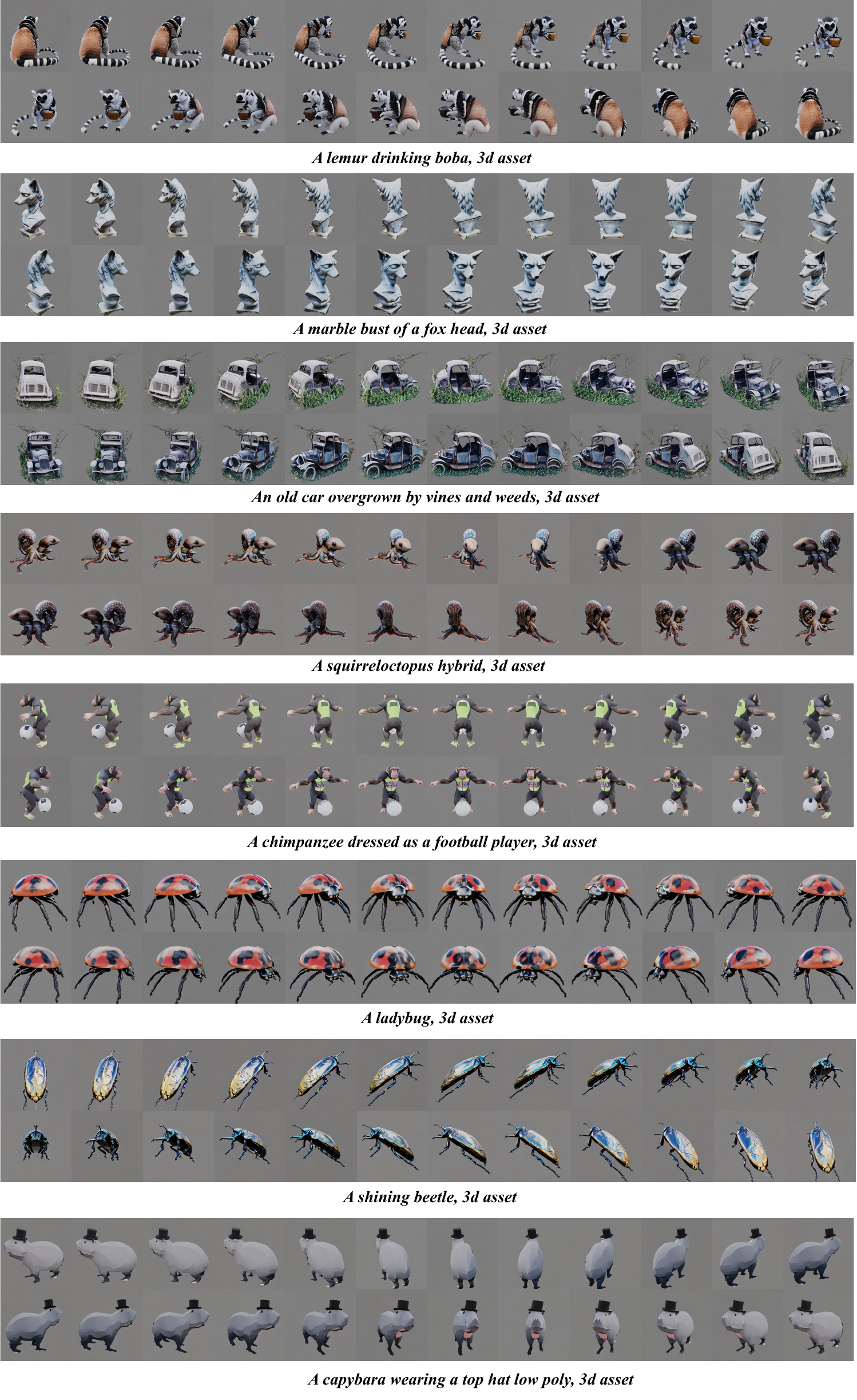}
  \vspace{-8mm}
  \caption{Visual results of Text-to-Multi-View image generation~(Part IX), prompts from DreamFusion.}
  \label{fig:qua9}
\end{figure}

Despite the limited data used for fine-tuning large-scale video generative models, the alignment between prompts and visual information in both video datasets and image datasets (consisting of 1-frame videos) remarkably enhances generalizability to open-vocabulary scenarios, enabling VideoMV to generate multi-view images beyond the limitations of training datasets. To further enhance qualitative visualization, we adopt highly abstract prompts previously employed in DreamFusion\cite{poole2022dreamfusion}. As illustrated in \cref{fig:qua1} to \cref{fig:qua9}, VideoMV consistently generates dense multi-view images based on abstract prompts, showing its ability to understand out-of-distribution data.

\newpage
\newpage
\section{Image-to-Multi-View Image Generation}

\begin{figure}[htbp]
  \centering
  \includegraphics[width=1.0\linewidth]{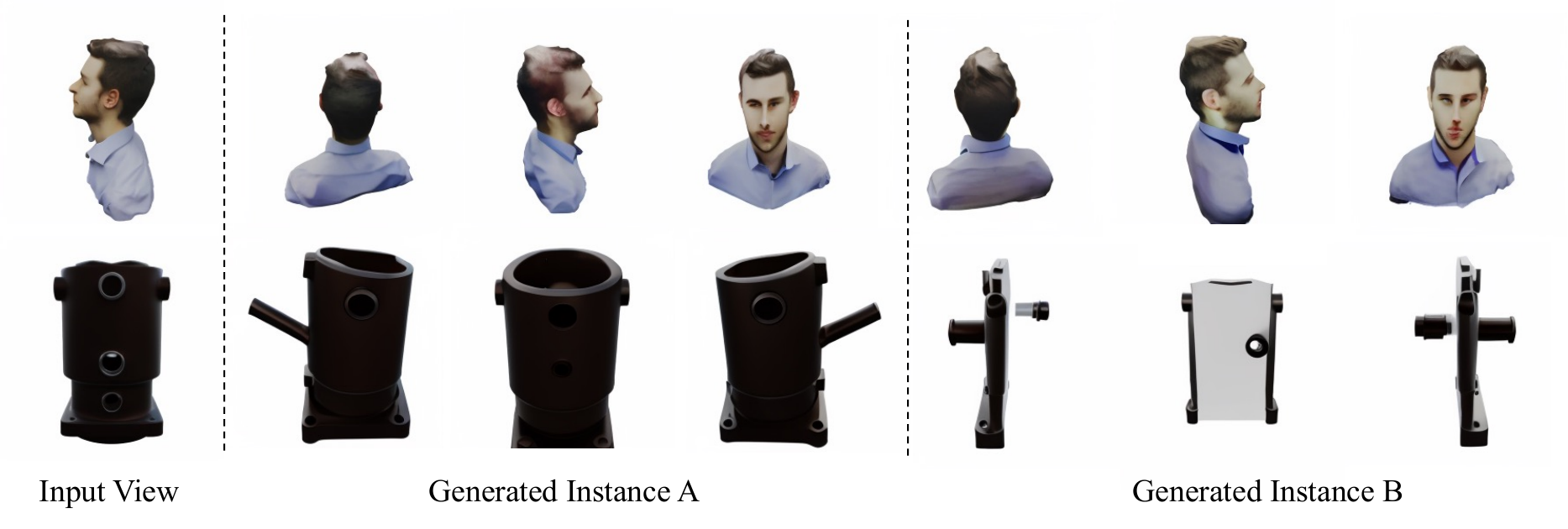}
  \caption{Visual various results of Image-to-Multi-View image generation.}
  \label{fig:img2mv_variance}
\end{figure}

Although the input image provides some guidance for dense pixel generation in multi-view scenarios, VideoMV is capable of generating various plausible results even from invisible angles. We present two typical examples in \cref{fig:img2mv_variance} to illustrate that VideoMV can produce diverse yet faithful outputs based on the given image input.

\vspace{-2.0 em}
\subsection{More Qualitative Results on Google Scanned Object}

\begin{figure}[htbp]
  \centering
  \includegraphics[width=1.0\linewidth]{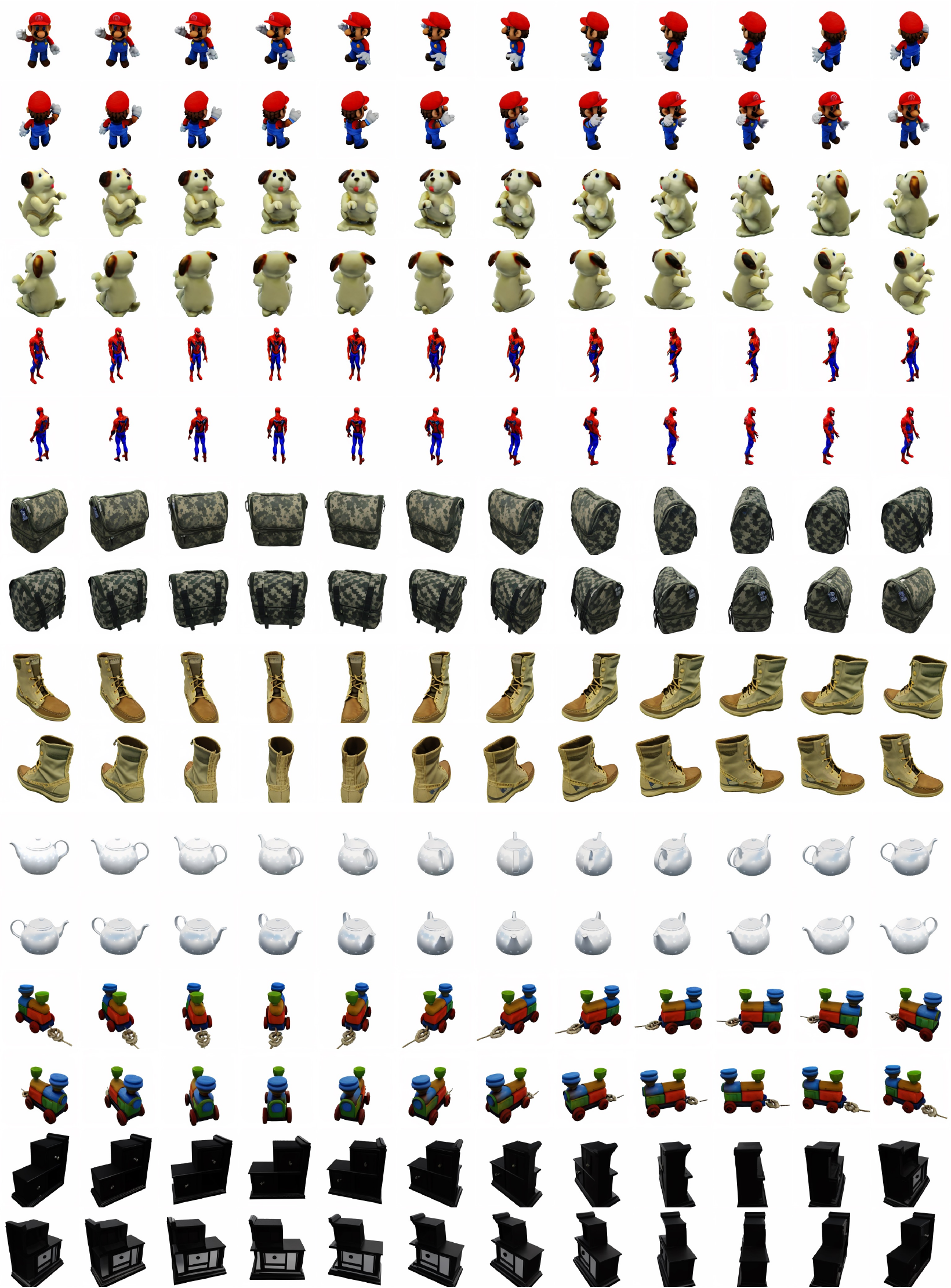}
  \caption{Visual results of Image-to-Multi-View image generation.}
  \label{fig:img2mv_vqua}
\end{figure}

Due to the page limitations in the main paper, we have included additional qualitative results of Google Scanned Objects in \cref{fig:img2mv_vqua} for a comprehensive analysis. Note that the first image serves as the input for VideoMV.

\subsection{More Qualitative Results on Web Images}

\begin{figure}[htbp]
  \centering
  \includegraphics[width=1.0\linewidth]{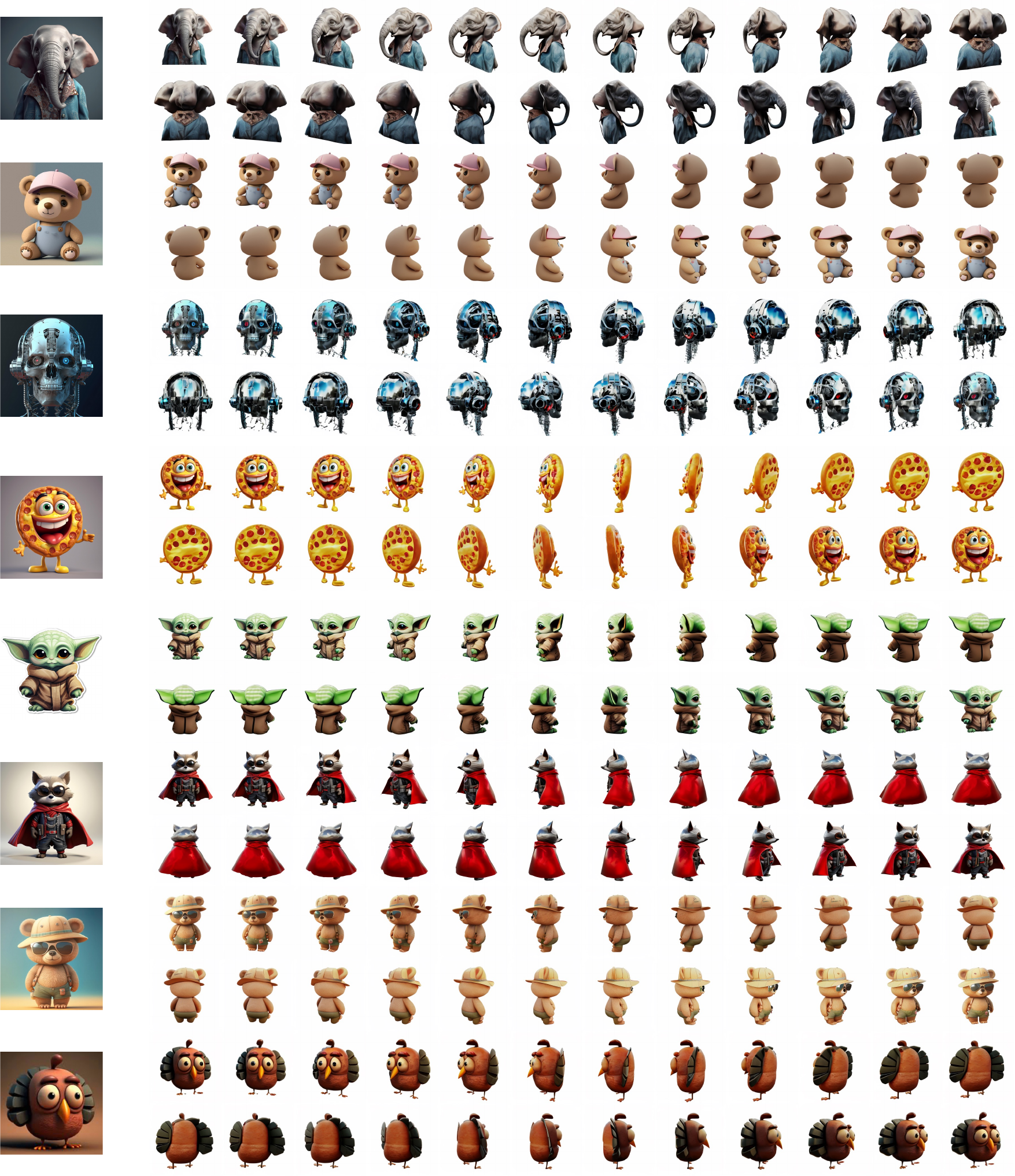}
  \caption{Visual results of Image-to-Multi-View image generation from web images.~(Part I)}
  \label{fig:img2mv_web_1}
\end{figure}

\begin{figure}[htbp]
  \centering
  \includegraphics[width=1.0\linewidth]{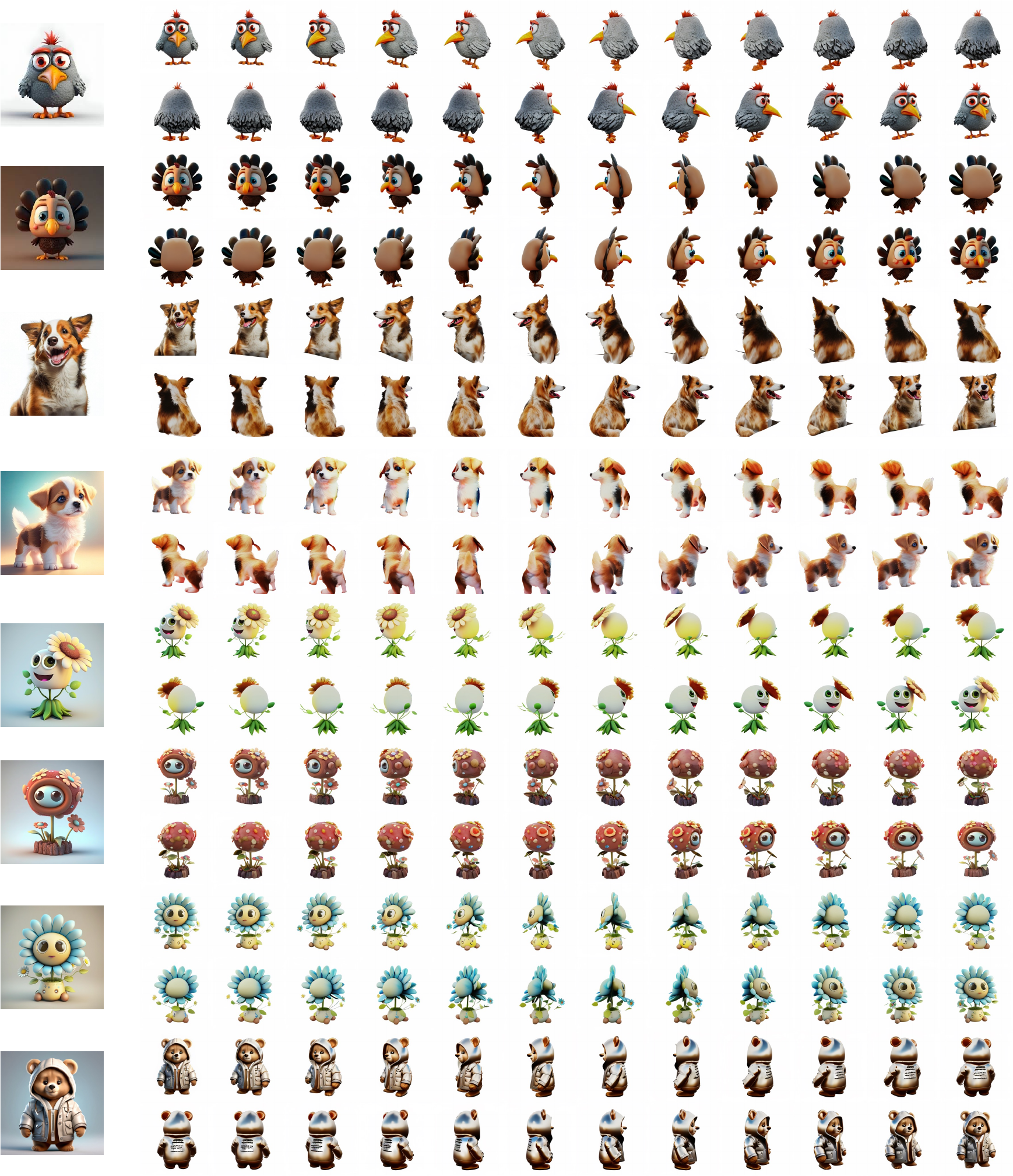}
  \caption{Visual results of Image-to-Multi-View image generation from web images.~(Part II)}
  \label{fig:img2mv_web_2}
\end{figure}

The VideoMV technique can also be applied to web images that lack underlying 3D models. In this study, we present a visualization of limited cases and encourage readers to experiment with our code on a wider range of web images. As illustrated in both \cref{fig:img2mv_web_1} and \cref{fig:img2mv_web_2}, VideoMV demonstrates its capability to generate feasible results using either generated or in-the-wild images.

\subsection{Numerical results on image-to-3D}

As a dense multi-view generative model, VideoMV aims to tackle the challenging task of synthesizing novel views with higher density and consistency based on a given prompt or single image. Unlike previous approaches\cite{liu2023zero1to3, liu2023syncdreamer}, we do not employ any reconstruction optimization in VideoMV. Instead, inspired by prior work\cite{wang2021neus}, we present relevant Volume IOU and Chamfer Distance metrics on the GSO dataset using the off-the-shelf MVS method, such as NeuS~\cite{wang2021neus}. As depicted in \cref{tab:supp_im3d}, VideoMV outperforms state-of-the-art methods in terms of Chamfer Distance and Volume IOU metrics, indicating that leveraging increased consistency in multi-view images for reconstruction can result in improved accuracy in 3D geometry.

\begin{table}
\begin{center}
    \caption{Quantitative results of image-to-3D on GSO~\cite{gso} dataset}
    \begin{tabular}{l c c }
        \toprule
         Method &  Chamfer Dist. & Volume IOU \\
        \hline
        \hline
        Zero123-XL & 0.0354 & 0.4846 \\
        SyncDreamer & 0.0278 & 0.5156 \\
        VideoMV & \textbf{ 0.0257} & \textbf{0.5228} \\
        \bottomrule
    \end{tabular}
\label{tab:supp_im3d}
\end{center}
\vspace{-2.0em}
\end{table}

\end{document}